\pdfoutput=1

\documentclass[11pt]{article}
\usepackage{acl}
\usepackage{booktabs,xcolor,siunitx}
\usepackage{colortbl} 
\definecolor{indomain}{RGB}{144,238,144} 
\definecolor{inndomain}{RGB}{253,127,57} 
\definecolor{outdomain}{RGB}{173,216,230} 
\definecolor{lightgray}{gray}{0.9}
\usepackage{times}
\usepackage{latexsym}
\usepackage{graphicx}

\usepackage{amsmath}
\usepackage{amssymb}
\usepackage{url}
\usepackage{booktabs}
\usepackage{ulem}
\usepackage{amsmath}
\usepackage{multirow}
\usepackage[T1]{fontenc}
\usepackage[utf8]{inputenc}
\usepackage{microtype}
\usepackage{inconsolata}

\title{Self-Refine Instruction-Tuning for Aligning Reasoning in Language Models}

\author{\textbf{Leonardo Ranaldi $^{(\dagger)}$, Andrè Freitas$^{(\dagger,*)}$} \\
	${(\dagger)}$ Idiap Research Institute, Martigny, Switzerland \\
 ${(*)}$Department of Computer Science, University of Manchester, UK \\
		{
  {\tt [name].[surname]@idiap.ch}
  } } 

\begin{document}
\maketitle
\begin{abstract}
The alignments of reasoning abilities between smaller and larger Language Models are largely conducted via Supervised Fine-Tuning (SFT) using demonstrations generated from robust Large Language Models (LLMs). Although these approaches deliver more performant models, they do not show sufficiently strong generalization ability as the training only relies on the provided demonstrations.

In this paper, we propose the \textit{Self-refine Instruction-tuning} method that elicits Smaller Language Models to self-refine their abilities.
Our approach is based on a two-stage process, where reasoning abilities are first transferred between LLMs and Small Language Models (SLMs) via Instruction-tuning on demonstrations provided by LLMs, and then the instructed models Self-refine their abilities through preference optimization strategies.

In particular, the second phase operates refinement heuristics based on the Direct Preference Optimization algorithm, where the SLMs are elicited to deliver a series of reasoning paths by automatically sampling the generated responses and providing rewards using ground truths from the LLMs.
Results obtained on commonsense and math reasoning tasks show that this approach significantly outperforms Instruction-tuning in both in-domain and out-domain scenarios, aligning the reasoning abilities of Smaller and Larger Language Models.

\end{abstract}

\section{Introduction}

Previous works have demonstrated that Chain-of-Thought (CoT) prompting can improve the Large Language Models (LLMs) \footnote{(e.g., with more than 60B parameters \cite{wei2023chainofthought})} capacity to perform complex reasoning tasks by decomposing a reasoning task into a sequence of intermediate steps \cite{wei2022emergent}, where the generation of multi-step controlled reasoning can improve results in commonsense \cite{bubeck2023sparks}, symbolic and mathematical \cite{gaur-saunshi-2023-reasoning,liu2023evaluating} reasoning datasets.

Since the size of LLMs represents an adoption barrier for many use cases and smaller models do not seem to have the same emergent reasoning abilities as LLMs, several state-of-the-art alignment approaches for solving mathematical problems have emerged, where Supervised Fine-Tuning (SFT) has been used to train Small Language Models (SLMs) using CoT annotations.
However, these annotations outline the intermediate reasoning steps for solving a given problem, which consists of a reasoning pathway generated by the LLM for the specific case. This phenomenon can lead to a relatively weak generalization capacity of tuned models that have a few and limited number of samples. Indeed, there are often multiple valid CoT annotations for the same question \cite{Cobbe2021TrainingVT,zhang2023interpretable}, which underlines the need for a more general CoT-based fine-tuning approach. 

In this paper, we propose \textit{Self-refine Instruction-tuning}, which is a method to enable CoT reasoning over SLMs.
Our approach starts by performing Instruction-tuning on SLMs via demonstrations delivered by LLMs and then applies preference optimization based on reinforcement learning (RL) heuristics to let the SLMs refine their abilities to solve a task in a step-wise manner.
Hence, proposing a teacher-student alignment method, we investigate the impact of transferring Chain-of-Thought reasoning abilities through the support of Demonstrations "taught" by LLMs to SLMs as a warm-up to the Self-refine process. 
Therefore, to reinforce the Instruction-tuning phase, we analyze whether preference optimization methods could strengthen students' step-wise reasoning abilities. 
\begin{figure*}[t]
\centering
    \includegraphics[width=0.93\textwidth]{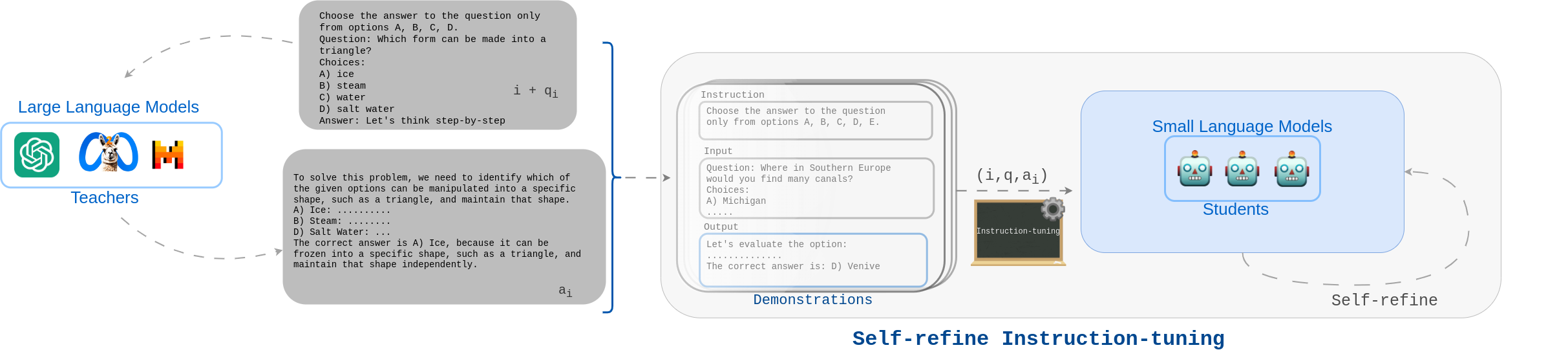}
    \caption{In Self-refine Instruction-tuning, the Demonstrations delivered by teacher models are used to align reasoning abilities in a teacher-student setting. Following the transference of step-wise reasoning knowledge via instruction tuning, the students Self-refine their abilities with the support of Direct Preference Optimization methods.}
    \label{fig:our_proposal}
\end{figure*}

Complementing the foundation work of \cite{wang-etal-2023-self-instruct,wang-etal-2023-democratizing}, we introduce Self-refinement based on reinforcement learning, and in contrast to \cite{uesato2022solving,luo2023wizardmath,luong2024reft,paul2024refiner}, we use an Instruction-tuning via Demonstrations approach \cite{ranaldi-freitas-2024-aligning} (i.e., a task-oriented specialization of Supervised Fine-Tuning) through which we instruct SLMs using Demonstrations delivered from different teachers prompted via a CoT mechanism. 

This leads to the target research questions, which are the focus of this paper: 

\textbf{RQ1:} How does Instruction-tuning via Demonstrations initialize the SLMs' reasoning abilities?

\textbf{RQ2:} What is the effect of the preference optimization algorithm on the alignment between teacher and student models?

\textbf{RQ3:} How much does the ability to solve tasks in a multi-step manner improve across different scenarios?

To answer these questions, we select three different SLMs: Llama2-7b, -13b \cite{touvron2023llama}, Mistral-7b \cite{jiang2023mistral}; and three LLMs Llama2-70b, Mixtral \cite{jiang2024mixtral} and GPT-3.5 \cite{openai2023gpt4}. 
In the teacher-student alignment phase, we use LLMs (teachers) to deliver Demonstrations at the core of the Instruction-tuning process (see Figure \ref{fig:our_proposal}) used to instruct SLMs (students). 
In the Self-refine phase, the students improve their step-wise reasoning abilities via Direct Preference Optimization (DPO) \cite{rafailov2023direct}. This allows the students to sample different reasoning paths and CoT Demonstrations and learn from them (Figure \ref{fig:our_proposal}). Moreover, differently from previous works, preferences are self-generated, and there is no need for a separately trained reward model as in the previous approaches \cite{ouyang2022training}.
We demonstrate the effectiveness of the proposed refinement technique in aligning teacher-student models (overcoming the differences highlighted by \citet{ranaldi-freitas-2024-aligning}) from the same family and in maximizing efficiency in in-domain and out-domain tasks.

Our contributions can be summarized as follows:
\begin{itemize}

\item We propose the Self-refined Instruction-tuning approach that is a task-oriented Supervised Fine-Tuning (SFT), which utilizes DPO heuristics to conduct a self-refinement process starting from instructed SLMs.

\item We analyze the impact of different configurations of Instruction-tuning on the SLMs before and after the Self-refining phase by conducting in-depth experiments on mathematical problems and common sense question-answering tasks using Demonstrations delivered by teacher of the same family (in-family) or not (out-family). Hence, we show the downstream functionalities in both scenarios.

\item Finally, we display the generalization abilities acquired via Self-refined Instruction-tuning through a systematic evaluation using Demonstrations provided by in-family and out-family teachers, both within in-domain and out-domain tasks.

\end{itemize}

\section{Method}
\label{sec:method}
To transfer the step-wise reasoning properties from Large Language Models (LLMs) to Small Language Models (SLMs), we propose \textit{Self-refine Instruction-tuning}, a two-step approach as shown in Figure \ref{fig:our_proposal}. In the first phase, there is a transfer of step-wise (CoT) reasoning via Instruction-tuning, where LLMs systematically generate Demonstrations which are used by SLMs to initialize their step-wise (CoT) alignment (Section \ref{sec:phase1}). In the second phase, the instructed SLMs Self-refine their internal CoT model via the preference optimization technique presented in Section \ref{sec:phase2}.

\subsection{Instruction-tuning Phase}
\label{sec:phase1}
A significant part of the state-of-the-art works employs standard Supervised Fine-Tuning (SFT) performed on annotations produced by a single LLM (Large Language Model) as a mechanism to improve SLMs. In our contribution, we take a step further and use Instruction-tuning, which is a task-oriented specialization of SFT (Supervised Fine-Tuning), in coordination with a teacher-student alignment approach (detailed in Appendix \ref{sec:Appendix_Instruction-tuning}).
In this phase, the SLM (student) is fine-tuned on a dataset produced by LLM (teacher) comprising a set of tuples in the form of \((i, q, a_i)\), where \(i\) represents a specific instruction, \(q\) is the input question (e.g., math-word problem), and \(a_i\) is the expected output and CoT answers generated from the teacher in response to the instruction and input. This setup is intended to transfer to the student models foundational problem-solving abilities, emphasizing the generation of outputs that conform to the provided instructions. The CoT answer \(a_i\) is articulated as:
\[a_i = [w_1, w_2, \ldots, w_{l-1}, w_l]\]
with \(l\) indicating the sequence length. At each timestep \(t\), the action \(w_t\) is derived from the policy \(\pi_\theta(\cdot | s_t)\), where \(w_t\) can be any token from the models vocabulary, and the state \(s_t\) encapsulates the concatenation of all previously generated tokens and the optional input \(x\) if provided. The state transition is defined as:
\[
s_{t+1} = \begin{cases} 
(x, i) & \text{if } t = 0 \\
[s_t, w_t] & \text{if } 1 \leq t \leq l
\end{cases}
\]

The Instruction-tuning loss function explicitly integrates the instruction \(i\), aligning the models' learning process with the instructional context. This loss function is formulated as:
\[
\mathcal{L}_{\text{inst}}(\theta) = -\mathbb{E}_{(i, q, a_i) \sim D} \left[\sum_{t=1}^{L} \log \pi_\theta(w_t | s_t, i) \right]
\]

Here, \(\pi_\theta\) is conditioned on both the state \(s_t\), the input \(q\), and the instruction \(i\), ensuring that the model prioritizes instruction compliance in its output generation. This methodological shift from SFT to Instruction-tuning underlines the principle of enhancing the models' ability to accurately interpret and execute complex instructions.

\subsection{Self-refinement Phase}
\label{sec:phase2}

In the second phase, the instructed SLMs (students) that have improved CoT properties via Instruction-tuning (Section \ref{sec:phase1}) self-refine these properties with the support of Direct Preference Optimization (DPO) \cite{rafailov2023direct}. This refinement can be conducted in an SFT style, relying exclusively on labeled preference data. The policy model, defined as \(\pi_\theta\), learns by repeatedly sampling the answers generated by teachers and students.

\paragraph{Direct Preference Optimization}
In the standard DPO approach \cite{rafailov2023direct}, a human annotator ranks the outputs from a reference policy, labeling winning and losing pairs $y_w = \pi_{inst}(x)$ and $y_l = \pi_{inst}(x)$. However, we propose an optimization step via Self-generated annotation by the students $\pi_{inst}$, which, after Instruction-tuning, should have more robust performances and reliably follow the demands of the questions.

For each Demonstration \((i, x, a_i)\), we prompt the students using the input $x = i + q$ ( or $x_{CoT} = x +  \texttt{"Let's think step by step"}$) (blue block in Figure \ref{fig:our_proposal}). Hence, for each instance within the Demonstrations we collect the \textbf{Answers} ($y_a = \pi_{\text{inst}}(x)$) that are the answers generated by the student given the input $x$, and the \textbf{CoT-Answers} ($y_{CoT} = \pi_{\text{inst}}(x_{CoT})$) are the answers that deliver CoT generated by the student elicited via CoT mechanism $x_{CoT}$.

In particular, assuming it is preferable for the model to generate responses that provide a CoT when elicited with $x_{CoT}$ and responses when prompted with $x$ just as the corresponding LLM teacher would do, we propose an alignment by exploiting DPO optimization. This aims to move the default style of our model (response generated by the student) towards the desired style (answers that deliver CoT). Different configurations are proposed depending on the desired result. Starting from the standard equation \ref{eq:1}:

\begin{equation}
\begin{aligned}
\mathcal{L}_{\texttt{DPO}}(\pi_{\theta}; \pi_{\text{inst}}) = -\mathbb{E}_{(x, y_w, y_l) \sim D} \\
\left[ \log \sigma (M(x, y_{w}, y_{l})) \right]
\end{aligned}
\label{eq:1}
\end{equation}

where $\sigma$ is the sigmoid function, and
\begin{equation}
 M(x, y_w, y_l) = \beta \log \frac{\pi_{\theta}(y_w|x)}{\pi_{\text{sf}t}(y_w|x)} - \beta \log \frac{\pi_{\theta}(y_l|x)}{\pi_{\text{sf}t}(y_l|x)}
\label{eq:M}
\end{equation}
where $\beta$ is a hyperparameter.

We propose the \textbf{Self-refine Instruction-tuning} that uses as optimization technique \textbf{\texttt{DPO$_{CoT}$}} (described in details in Appendix \ref{sec:APP_DPO} in Equation \ref{eq:DPO_CoT}. In particular, in \textbf{\texttt{DPO$_{CoT}$}} the answers that deliver a CoT response which is self-generated from the students are referred to as the preferred response. 

\section{Experimental Setup}
\label{sec:Experimental_Setup}
In order to evaluate the proposed model, we use both commonsense and mathematical reasoning tasks (introduced in Section \ref{sec:data}) that are generally used to assess the step-wise inference properties of Large Language Models (LLMs). Regarding the Self-refine Instruction-tuning on the Small Language Models (SLMs), we use the approach presented in Section \ref{sec:Teach_smallLLMs}.

\subsection{Tasks \& Datasets}
\label{sec:data}

In this paper, we selected different tasks that focus on reasoning tasks: 

\paragraph{Commonsense Task} We adopt two benchmarks to evaluate commonsense reasoning: CommonSenseQA \cite{talmor-etal-2019-commonsenseqa} (CSQA) and OpenBookQA \cite{mihaylov2018suit} (OBQA) are two multi-choice commonsense question-answering tasks. 

\paragraph{Physical \& Social Interaction Task} We adopt two benchmarks to evaluate reasoning in the context of everyday situations, aiming to establish the most reasonable solution: Interaction Question Answering (PIQA) \cite{bisk2019piqa} and Social Interaction Question Answering (SIQA) \cite{sap-etal-2019-social}, which emphasizes people's actions and social implications.

\paragraph{Mathematical Task} We use two math word problem benchmarks to evaluate the models of mathematical reasoning. MultiArith \cite{roy-roth-2015-solving} covers a set of multi-step arithmetic reasoning tasks, while GSM8k \cite{Cobbe2021TrainingVT} covers a set of primary school-level mathematical problems. 

\paragraph{Additional benchmarks} Finally, to evaluate the adaptability of our proposal, we conduct further analysis on two additional evaluation benchmarks: MATH \cite{hendrycks2021measuring}, and MMLU \cite{hendrycks2020measuring}.

\paragraph{Datasets}
Since the test split is not prescribed for all the benchmarks, we adopt the following strategy: for SIQA, PIQA, CSQA, and OBQA, we use 4000 examples with equally distributed target classes as training data and the validation versions found on huggingface as test data, while for GSM8K and MultiArith we use the full huggingface datasets. In Table \ref{tab:dataset_summary}, we report the descriptive statistics and splitting ratios, while in Table \ref{tab:examples_benchmarks}, we report one example for each benchmark. The supporting datasets are publicly accessible as described in Table \ref{tab:versions_data_HF}.

\subsection{Self-refine Instruction-tuning Pipeline}
\label{sec:Teach_smallLLMs}

The Self-refine Instruction-tuning comprises the annotation process conducted by the LLMs teachers that are prompted in the zero-shot scenario (as shown in Table \ref{tab:esempio_input}), as explained in Appendix \ref{sec:Appendix_Instruction-tuning}.
We selected Llama-2-70 \cite{touvron2023llama}, Mixtral7x8 \cite{jiang2024mixtral} and GPT-3.5 \cite{openai2023gpt4} as LLMs (teachers) and Llama2-7, -13 \cite{touvron2023llama} and Mistral-7 \cite{jiang2023mistral} SMLs (students) models. 

Hence, the students models are tuned, as proposed in \cite{alpaca} and evaluated with probing pipelines (detailed in Section \ref{sec:Evaluation}). The students are instructed via Demonstrations that contain the answers generated by the teachers, as explained in Section \ref{sec:phase1}. 
Downstream of the teacher-student CoT transference process, the optimization technique (proposed in Section \ref{sec:phase2} and detailed in Appendix \ref{sec:APP_DPO}) is employed to improve alignment and self-refine the quality of the generation.

\begin{figure*}[t]
\centering
         \begin{minipage}{0.3\linewidth}
     \centering
     \includegraphics[width=\linewidth]{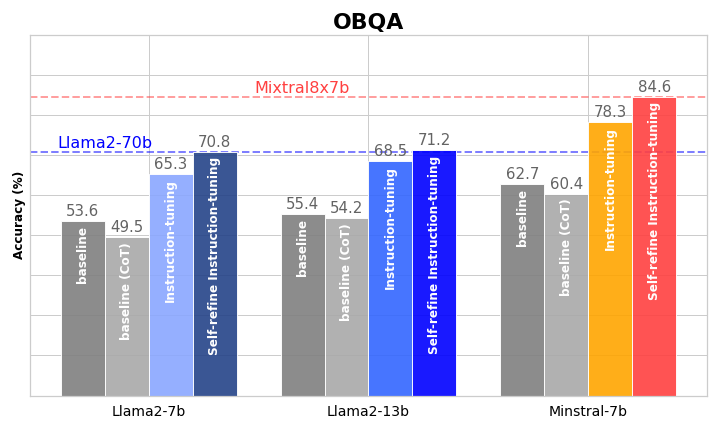}
   \end{minipage}
            \begin{minipage}{0.3\linewidth}
     \centering
     \includegraphics[width=\linewidth]{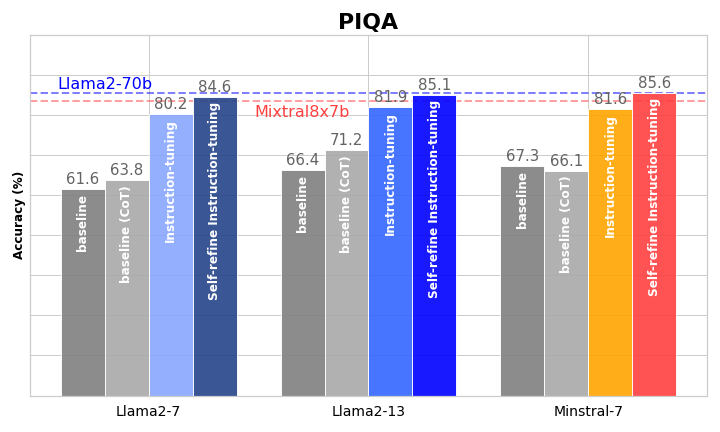}
   \end{minipage}
    \begin{minipage}{0.3\linewidth}
     \centering
     \includegraphics[width=\linewidth]{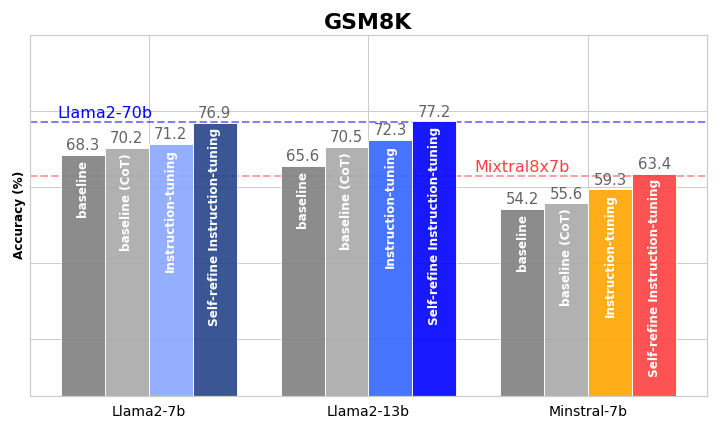}
   \end{minipage}
         \begin{minipage}{0.3\linewidth}
     \centering
     \includegraphics[width=\linewidth]{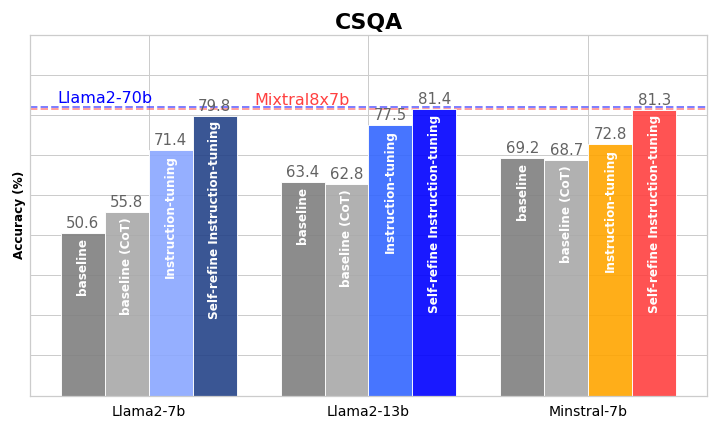}
   \end{minipage}
            \begin{minipage}{0.3\linewidth}
     \centering
     \includegraphics[width=\linewidth]{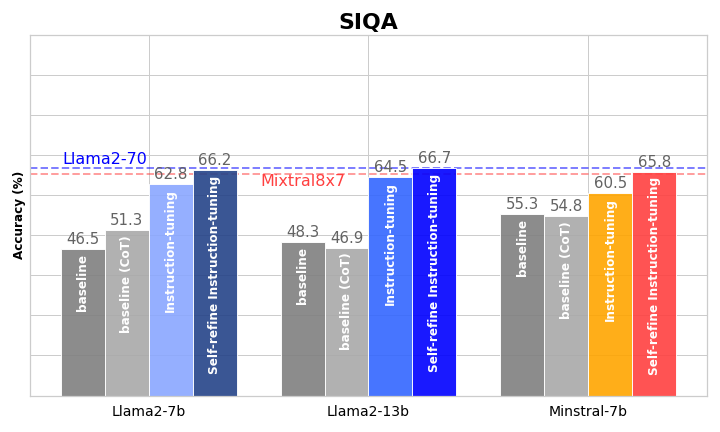}
   \end{minipage}   
    \begin{minipage}{0.3\linewidth}
     \centering
     \includegraphics[width=\linewidth]{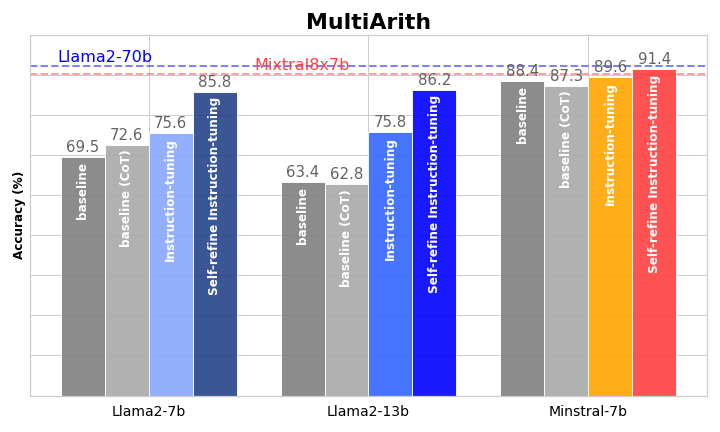}
   \end{minipage}

        \begin{minipage}{0.64\linewidth}
     \centering
     \includegraphics[width=\linewidth]{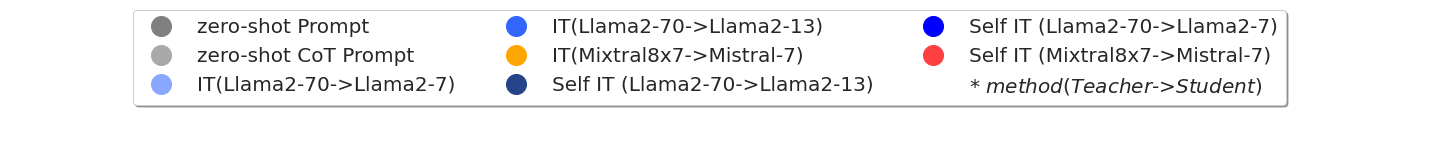}
   \end{minipage}

   \caption{Accuracies (\%) on benchmarks (Section \ref{sec:data}) before Instruction-tuning (i.e., \texttt{Baselines} and \texttt{Baseline CoT}), after Instruction-tuning (IT) performed on Demonstrations delivering CoT and finally behind the \texttt{Self-refine Instruction-tuning} phase (Self IT). In particular, the models were instructed via Demonstrations delivered by in-family LLMs (as described in the legend, we use the notation \textit{method(Teacher->Student)}). } 
   \label{fig:performances_in_family}

\end{figure*}

\subsubsection{Models Setup}
\label{sec:Models_Setup}

We conduct the Self-refined Instruction-tuning in two different phases. Firstly, we start with Instruction-tuning phase using QLoRA \citet{dettmers2023qlora}. This approach allows Instruction-tuning to be performed while reducing memory usage. In particular, \citet{dettmers2023qlora} propose several techniques for tuning models with many parameters on GPUs with limited resources while preserving 16-bit tuning performance.
We follow the training approach proposed in \cite{alpaca}, setting four training epochs using a learning rate of 2e-5 with a 1e-4 weight decay. We use the cosine learning rate scheduler with a warm-up ratio of 0.03. 
Furthermore, we conduct the Self-refine phase following the approach proposed in \cite{rafailov2023direct}. In particular, we use the huggingface $DPO_{trainer}$ to support its reproducibility. We follow the parameters proposed in \cite{rafailov2023direct}. Hence, for the DPO policy, our work employs a learning rate of 1e-6, $\beta$ set at $0.1$, and a warm-up step count of 100. The batch size is configured to 128. The optimization process is capped at a maximum of 1000 steps, where we save the checkpoint corresponding to the lowest loss on the validation set. The experiments were conducted on a workstation equipped with four Nvidia RTX A6000 with 48GB of VRAM.

\begin{figure*}[t]
\centering
         \begin{minipage}{0.3\linewidth}
     \centering
     \includegraphics[width=\linewidth]{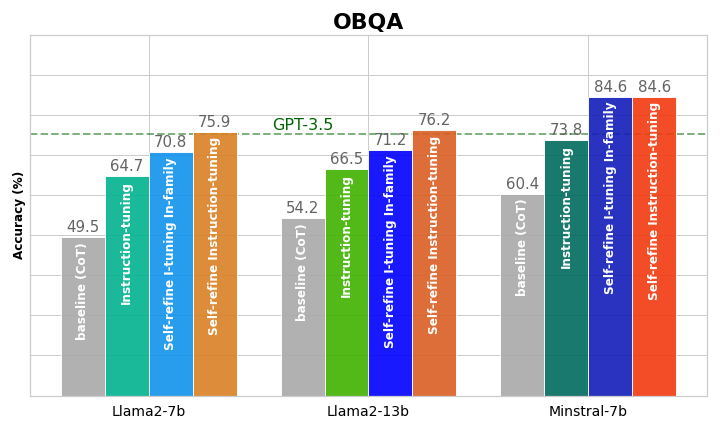}
   \end{minipage}
            \begin{minipage}{0.3\linewidth}
     \centering
     \includegraphics[width=\linewidth]{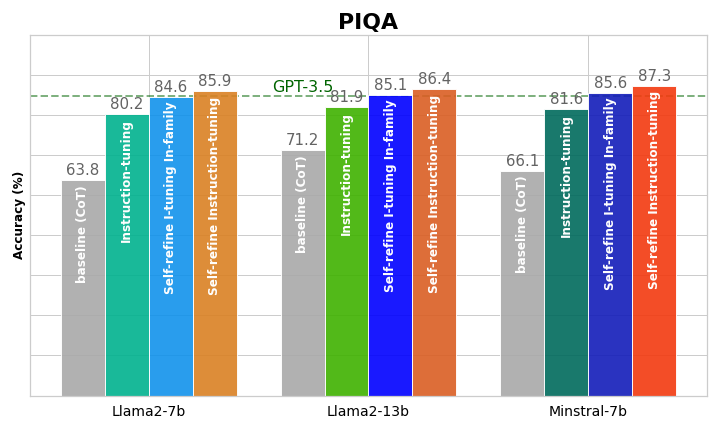}
   \end{minipage}
    \begin{minipage}{0.3\linewidth}
     \centering
     \includegraphics[width=\linewidth]{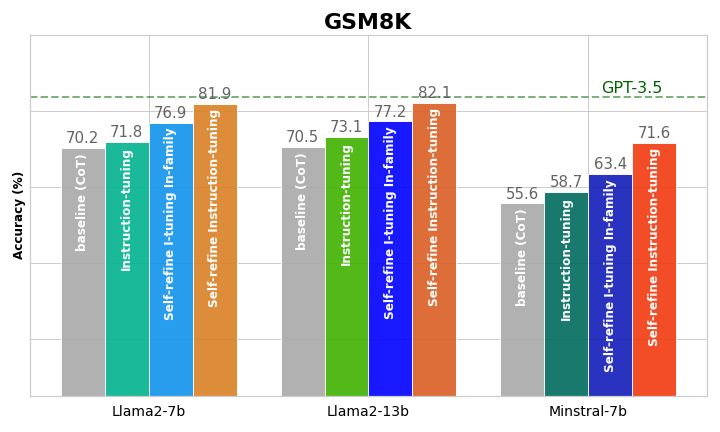}
   \end{minipage} 
         \begin{minipage}{0.3\linewidth}
     \centering
     \includegraphics[width=\linewidth]{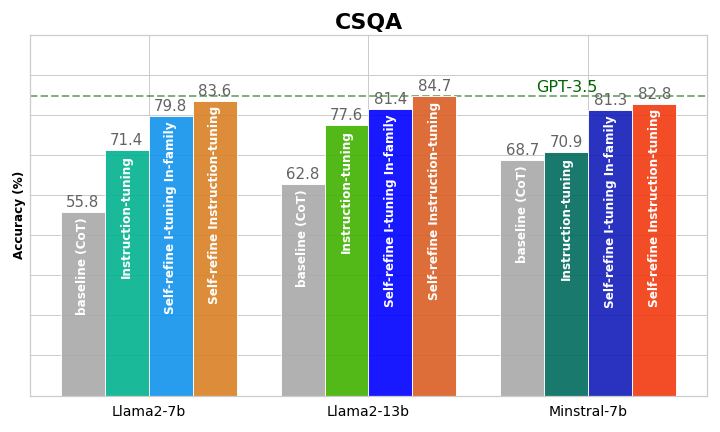}
   \end{minipage}
            \begin{minipage}{0.3\linewidth}
     \centering
     \includegraphics[width=\linewidth]{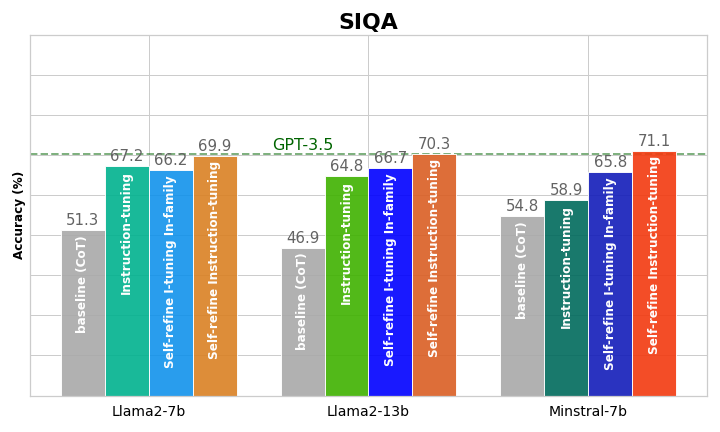}
   \end{minipage}
    \begin{minipage}{0.3\linewidth}
     \centering
     \includegraphics[width=\linewidth]{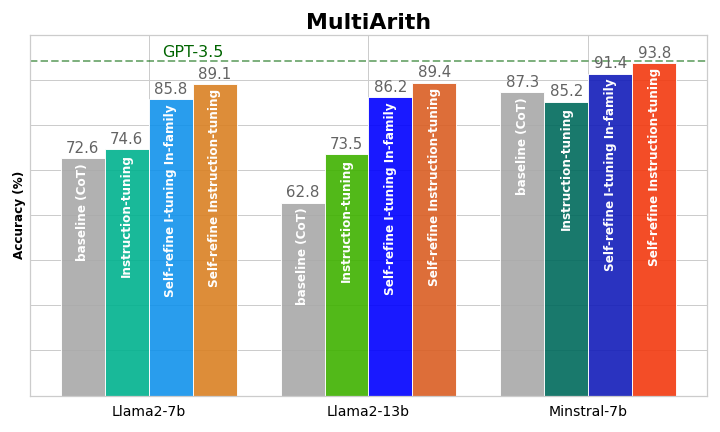}
   \end{minipage}

        \begin{minipage}{0.64\linewidth}
     \centering
     \includegraphics[width=\linewidth]{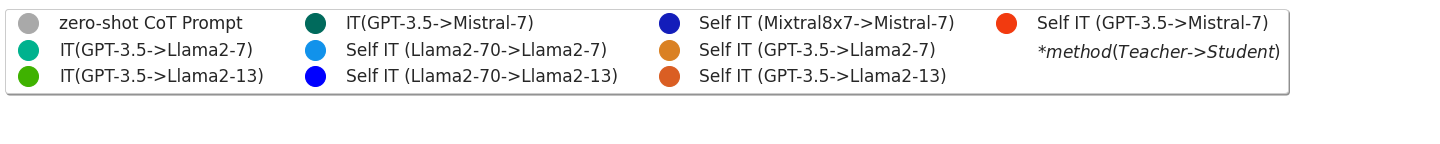}
   \end{minipage}

   \caption{Accuracies (\%) on benchmarks (Section \ref{sec:data}) before Instruction-tuning ( Baseline CoT), behind first phase performed on Demonstrations delivering CoT (i.e., Instruction-tuned (IT)) and finally behind the Self-refine phase (i.e., Self-refine IT). In particular, the models were instructed via Demonstrations delivered by out-family LLMs (as described in the legend, we use the notation \textit{method(Teacher->Student)}).} 
   \label{fig:performances_out_family}

\end{figure*}

\subsection{Evaluation}
\label{sec:Evaluation}
The most commonly used evaluation methods for question-answering tasks are language-model probing, in which the option with the highest probability is selected \cite{brown2020language}, and multiple-choice probing, in which the models are asked to commit to an answer. The evaluation in the first case is performed with a function taking the \textit{argmax} and, in the second case, with a direct string matching. The second method is more widely used in recent evaluations as it can be inclusive to the larger GPT family models\cite{openai2023gpt4}, where probability values are not readily accessible. In the experiments, we chose the latter to have a comparable and scalable pipeline (Details provided in Appendix \ref{sec:parameters}). Finally, string matching is performed between the generated outputs and the target choice to evaluate the percentages of the correct answers. 

\section{Results \& Discussion}

The \textit{Self-refine Instruction-tuning} improves the alignment between Large Language Models (LLMs) and Small (SLMs) in both in-family and out-family settings. These conclusions can be observed in Figure \ref{fig:performances_in_family} and Figure \ref{fig:performances_out_family}, which reports the downstream accuracies without tuning (see the Baselines), with only the Instruction-tuning phase on Demonstrations and after the Self-refine phase. As discussed in Section \ref{sec:res_Instruction_tuning}, the models with only Instruction-tuning on Demonstrations (generated by LLMs) transfers the reasoning properties in a marginal way (see Instruction-tuned in Figures \ref{fig:performances_in_family}). 

However, although teacher-student alignment via Instruction-tuning produces better students, an improved alignment is achieved through the Self-refine phase, as discussed in \ref{sec:res_Self_refine}. In particular, the 'Self-refine Instruction-tuning' bars in Figure \ref{fig:performances_in_family} show that the students self-refined outperformed the students tuned only with Instruction-tuning ('Instruction-tuning' bars on Figure \ref{fig:performances_in_family}). Furthermore, the alignment via Demonstrations generated by teachers outside the same family (out-family) delivers more robust students (see Figure \ref{fig:performances_out_family} the Self-refine Instruction-tuning and (in-family) bars).

Finally, students models behind the self-refine phase outperformed others in both in-domain and out-domain tasks (discussed in Section \ref{sec:res_in_domain_vs_out_domain}). Hence, the self-refine mechanism effectively aligns teacher-student capabilities in out-domain tasks by enhancing performance even in the presence of fewer Demonstrations (Section \ref{sec:res_ablation}).

\subsection{The Instruction-tuning alignment}
\label{sec:res_Instruction_tuning}
Instruction-tuning led by Larger Language Models (teachers models), which are able to deliver multi-step reasoned answers, induces this property within Smaller Language Models (students models). This can be seen in the experiments in Figure \ref{fig:performances_in_family}, Figure \ref{fig:performances_out_family} and additional evaluations in Appendix \ref{app:additionals}. The student models behind instruction-tuning on demonstrations produced by teacher models outperformed the baselines of the proposed benchmarks.

While one can observe consistent improvements in performance across the board, there are moderate variations across models and tasks. 
The teacher models that generate Demonstrations stem from different families and perform differently, as shown in Table \ref{tab:evaluations_teachers}. The consequence of this phenomenon can be seen in Figure \ref{fig:performances_in_family} and Figure \ref{fig:performances_out_family} (horizontal lines that are the reported performance of the teachers and bars 'Instruction-tuning' that are the performance of the students). Therefore, the teacher-student alignment is not complete as there is a gap between the performances of the teachers and the students tuned via Instruction-tuning (only phase presented in Section \ref{sec:phase1}). In addition, it is possible to differentiate between in-family and out-family alignment. In the in-family, where students are instructed with Demonstrations delivered by the teachers of the same family, performances vary from 6.3 points on average in question-answering (QA) tasks and 8.2 points on average in math word problems (MWP) tasks. Meanwhile, in the out-family alignment, the performances vary by 8.5 on the QA and 8.7 on the MWP.

Hence, to improve the alignment both in-family and consistently out-family, we have proposed an optimization technique based on a self-refinement approach (introduced in Section \ref{sec:phase2}), the results of which we discuss in Section \ref{sec:res_Self_refine}.

\begin{table*}[t]
\small
\centering
\begin{tabular}{l|c|ccccccc}
\toprule
\textbf{Trained on} & \textbf{Teacher} & \multicolumn{6}{c}{\textbf{Evaluated on}} \\
\cmidrule{3-8}
& & \textbf{OBQA} & \textbf{CSQA} & \textbf{PIQA} & \textbf{SIQA} & \textbf{GMS8K} & \textbf{MultiArith} \\
\midrule
{Baseline} & - & 53.6{\tiny $\pm .2$} & 50.6{\tiny $\pm .4$} & 61.6{\tiny $\pm .1$} & 46.5{\tiny $\pm .3$} & 68.2{\tiny $\pm .5$} & 69.5{\tiny $\pm .2$} \\
{Baseline CoT} & - & 49.5{\tiny $\pm .4$} & 55.8{\tiny $\pm .3$} & 63.8{\tiny $\pm .1$} & 51.3{\tiny $\pm .5$} & 71.3{\tiny $\pm .2$} & 72.6{\tiny $\pm .4$} \\
\hline
\multirow{3}{*}{\textbf{OBQA}} &  \texttt{Instruction-tuning} & \cellcolor{inndomain}65.3{\tiny $\pm .3$} & \cellcolor{indomain}65.4{\tiny $\pm .2$} & \cellcolor{indomain}66.3{\tiny $\pm .4$} & \cellcolor{indomain}59.2{\tiny $\pm .2$} & \cellcolor{outdomain}61.4{\tiny $\pm .2$} & \cellcolor{outdomain}60.2{\tiny $\pm .3$} \\
& \texttt{+ Self-refine} & \cellcolor{inndomain}\textbf{70.8}{\tiny $\pm .3$} & \cellcolor{indomain}73.2{\tiny $\pm .2$} & 
\cellcolor{indomain}75.3{\tiny $\pm .1$} & \cellcolor{indomain}62.6{\tiny $\pm .3$} & \cellcolor{outdomain}68.7{\tiny $\pm .4$} & \cellcolor{outdomain}69.8{\tiny $\pm .3$} \\
& \texttt{Cross Self-refine} & - & 78.4{\tiny $\pm .1$} & 
78.3{\tiny $\pm .5$} & 64.5{\tiny $\pm .3$} & 74.4{\tiny $\pm .4$} & 83.2{\tiny $\pm .2$} \\

\hline

\multirow{3}{*}{\textbf{CSQA}}
 & \texttt{Instruction-tuning} & \cellcolor{indomain}57.8{\tiny $\pm .1$} & \cellcolor{inndomain}71.4{\tiny $\pm .3$} & \cellcolor{indomain}65.5{\tiny $\pm .4$} & \cellcolor{indomain}61.8{\tiny $\pm .2$} & \cellcolor{outdomain}60.1{\tiny $\pm .5$} & \cellcolor{outdomain}59.3{\tiny $\pm .1$} \\
 &\texttt{+ Self-refine} & \cellcolor{indomain}69.5{\tiny $\pm .5$} & \cellcolor{inndomain}\textbf{79.8}{\tiny $\pm .3$} & \cellcolor{indomain}74.2{\tiny $\pm .1$} & \cellcolor{indomain}66.3{\tiny $\pm .2$} & \cellcolor{outdomain}61.2{\tiny $\pm .3$} & \cellcolor{outdomain}60.3{\tiny $\pm .3$} \\
 &\texttt{Cross Self-refine} & 68.7{\tiny $\pm .4$} & - & 78.4{\tiny $\pm .2$} & 64.1{\tiny $\pm .3$} & 72.1{\tiny $\pm .4$} & 73.4{\tiny $\pm .2$} \\
\hline

\multirow{3}{*}{\textbf{PIQA}} & \texttt{Instruction-tuning} & \cellcolor{indomain}56.9{\tiny $\pm .1$} & \cellcolor{indomain} 64.3{\tiny $\pm .2$} & \cellcolor{inndomain} 80.2{\tiny $\pm .3$} & \cellcolor{indomain} 57.3{\tiny $\pm .3$} & \cellcolor{outdomain} 58.3{\tiny $\pm .1$} & \cellcolor{outdomain} 59.1{\tiny $\pm .3$} \\
 & \texttt{+ Self-refine} & \cellcolor{indomain}68.2{\tiny $\pm .4$} & \cellcolor{indomain} 67.3{\tiny $\pm .5$} & \cellcolor{inndomain}\textbf{84.6}{\tiny $\pm .3$} & \cellcolor{indomain} 63.4{\tiny $\pm .2$} & \cellcolor{outdomain} 67.8{\tiny $\pm .1$} & \cellcolor{outdomain} 66.9{\tiny $\pm .3$} \\
 & \texttt{Cross Self-refine} & 68.2{\tiny $\pm .3$} &  71.3{\tiny $\pm .3$} &  - &  64.2{\tiny $\pm .1$} &  68.7{\tiny $\pm .4$} &  67.6{\tiny $\pm .1$} \\
\hline

\multirow{3}{*}{\textbf{SIQA}} & \texttt{Instruction-tuning} & \cellcolor{indomain}58.9{\tiny $\pm .2$} & \cellcolor{indomain} 62.8{\tiny $\pm .5$} & \cellcolor{indomain} 63.2{\tiny $\pm .1$} & \cellcolor{inndomain} 62.8{\tiny $\pm .3$} & \cellcolor{outdomain} 59.6{\tiny $\pm .1$} & \cellcolor{outdomain} 60.2{\tiny $\pm .3$} \\

& \texttt{+ Self-refine} & \cellcolor{indomain} 68.3{\tiny $\pm .3$} & \cellcolor{indomain} 68.5{\tiny $\pm .2$} & \cellcolor{indomain} 78.3{\tiny $\pm .3$} & \cellcolor{inndomain}\textbf{66.2}{\tiny $\pm .4$} & \cellcolor{outdomain}61.3{\tiny $\pm .5$} & \cellcolor{outdomain}60.9{\tiny $\pm .4$} \\
& \texttt{Cross Self-refine} &  69.4{\tiny $\pm .2$} &  68.5{\tiny $\pm .2$} &  77.9{\tiny $\pm .3$} & - & 65.1{\tiny $\pm .3$} & 64.7{\tiny $\pm .2$} \\
\hline

\multirow{3}{*}{\textbf{GSM8K}} & \texttt{Instruction-tuning} & \cellcolor{outdomain}53.2{\tiny $\pm .4$} & \cellcolor{outdomain}54.9{\tiny $\pm .5$} & \cellcolor{outdomain}63.7{\tiny $\pm .1$} & \cellcolor{outdomain}52.5{\tiny $\pm .2$} & \cellcolor{inndomain}71.2{\tiny $\pm .3$} & \cellcolor{indomain}70.3{\tiny $\pm .2$} \\
& \texttt{+ Self-refine} & \cellcolor{outdomain}58.6{\tiny $\pm .3$} & \cellcolor{outdomain}61.7{\tiny $\pm .4$} & \cellcolor{outdomain}62.3{\tiny $\pm .2$} & \cellcolor{outdomain}52.4{\tiny $\pm .3$} & \cellcolor{inndomain}\textbf{76.9}{\tiny $\pm .1$} & \cellcolor{indomain}74.3{\tiny $\pm .2$} \\
& \texttt{Cross Self-refine} & 64.6{\tiny $\pm .5$} & 64.3{\tiny $\pm .2$} & 77.6{\tiny $\pm .4$} & 60.3{\tiny $\pm .2$} & - & 75.3{\tiny $\pm .3$} \\
\hline

\multirow{3}{*}{\textbf{MultiArith}} & \texttt{Instruction-tuning} & \cellcolor{outdomain}53.6{\tiny $\pm .2$} & \cellcolor{outdomain}55.7{\tiny $\pm .3$} & \cellcolor{outdomain}53.8{\tiny $\pm .3$} & \cellcolor{outdomain}51.5{\tiny $\pm .3$} & \cellcolor{indomain}69.3{\tiny $\pm .1$} & \cellcolor{inndomain}75.6{\tiny $\pm .2$} \\
& \texttt{+ Self-refine} & \cellcolor{outdomain}59.1{\tiny $\pm .2$} & \cellcolor{outdomain}63.2{\tiny $\pm .5$} & \cellcolor{outdomain}58.3{\tiny $\pm .3$} & \cellcolor{outdomain}58.6{\tiny $\pm .1$} & \cellcolor{indomain}70.2{\tiny $\pm .4$} & \cellcolor{inndomain}\textbf{85.8}{\tiny $\pm .2$} \\
& \texttt{Cross Self-refine} & 65.3{\tiny $\pm .4$} & 61.3{\tiny $\pm .1$} & 62.1{\tiny $\pm .2$} & 60.7{\tiny $\pm .5$} & 73.4{\tiny $\pm .3$} & - \\
\bottomrule
\end{tabular}
\caption{Evaluation of Llama-2-7 Instruction-tuned (\texttt{Instruction-tuned}) and with completely Self-refine Instruction-tuning (\texttt{+ Self-refine Instruction-tuned}) on Demonstrations using different test sets. We evaluate in-domain (QA vs QA) and out-domain (QA vs math-word problem) benchmarks. "Baselines" are referred to the non-instructed model. Results colored in green indicate the in-domain benchmark, blue the out-domain benchmark, and orange the same benchmark on which perform the evaluation phase. Moreover, we propose Self-refine Instruction-tuning in cross-setting scenario where we optimize the model on the training set related to the evaluated task.}
\label{tab:generalization_results_llama_7}
\end{table*}

\subsection{The Self-refine Impact}
\label{sec:res_Self_refine}
The Self-refine process enables complete in-family student-teacher alignment by consistently increasing performance in out-family settings and improving the qualities of generated answers. The results obtained in Figure \ref{fig:performances_in_family} show that the students (SLMs instructed with Self-refine Instruction-tuning) outperform the non-self-refined students and perform comparably to their teachers. The same behaviour can be observed from the out-family setting shown in Figure \ref{fig:performances_out_family}. In particular, the teacher \texttt{GPT-3.5} showed a more robust baseline performance (Table \ref{tab:evaluations_teachers}). Although Instruction-tuning alone transfers some of the abilities to the student models, they were significantly lower when compared to the out-family teacher models. In contrast, the teacher-student performances significantly converged after the self-refine phase, leading to the alignment completion. Finally, a positive impact can also be observed on the quality of students' generations, as shown in the additional experiment discussed in Appendix \ref{app:quality}.

The performances appear completely aligned, but the students were tested only for in-domain tasks. The proposed approach could cause students to over-specialize in in-domain tasks, running the risk of losing the ability to solve out-domain tasks. For this reason, we performed a set of assessments evaluating students on in-domain and out-domain tasks and discussed the results in Section \ref{sec:res_in_domain_vs_out_domain}.

\subsection{In-Domain and Out-Domain}
\label{sec:res_in_domain_vs_out_domain}
The Self-refine Instruction-tuning approach complements student-teacher alignment and improves students' generalization abilities in out-domain tasks. These results can be observed in Table \ref{tab:generalization_results_llama_7} with \texttt{Llama2-7} as students and \texttt{Llama2-70} as teachers (in Appendix Table \ref{tab:generalization_results_llama_13} with Llama2-13 Table \ref{tab:generalization_results_mistral} with Mistral-7). In particular, behind the evaluations performed on in-domain and out-domain tasks, the students Self-refine Instruction-tuned outperform the baselines and the Instruction-tuned models.
Furthermore, to observe the impact of the optimization phase (introduced in Section \ref{sec:phase2}) on the downstream performance, we conducted a further experiment by fixing the Instruction-tuning phase and switching the Self-refine ones across different evaluation tasks (e.g., we instructed a student on OBQA and then optimized via self-refine approach on CSQA). As shown in lines \texttt{Cross Self-refine} of Table \ref{tab:generalization_results_llama_7}, students warmed up on tasks other than those they are optimized, outperformed the others, and obtained similar performances to those obtained from in-domain models. This shows that optimization positively impacts the alignment of generalization abilities in out-domain tasks.
Finally, following evaluations in out-domain tasks and across scenarios, we evaluate the performance of the proposed approach by reducing the number of demonstrations available for alignment in Section \ref{sec:res_ablation}.

\subsection{Low-resource Optimization}
\label{sec:res_ablation}
Self-refine Instruction-tuning achieves sustainable performances in low-resource settings. In fact, in Figure \ref{fig:performances_increasing}, it is possible to observe that the performance achieved by the self-refined students consistently outperforms that of the non-self-refined students (where only phase 1 described in Section \ref{sec:phase1} was performed) (technical details on the breakdown can be found in Appendix \ref{sec:experimental_details_splitting}). 
Although it emerges that only the optimization process via DPO is more performant than the instruction-tuning process alone, the combination of the two phases achieves the best results in both in-family and out-family alignment in each proposed splitting that are described in Appendix \ref{sec:experimental_details_splitting}.

\section{Related Work}

\subsection{Multi-step Reasoning}

Previous works focus on Chain-of-Thought (CoT) prompting techniques, studying the impact of prompting design and engineering, proposing specialized interventions to improve CoT generalization and fine-grained multi-step reasoning properties \cite{wei2022emergent,fu2023complexitybased}.

On the prompting design side, \citet{gao2023pal} proposed using Python programs as a CoT prompt, demonstrating more accurate reasoning steps and significant improvements behind CoT prompting \cite{wei2022emergent}. \citet{zhou2023solving} introduced a code generation approach to verify the intermediate reasoning step \cite{openai2023gpt4}. 

In parallel, there have been improvements in the accessibility of lower-parameter versions of Large Language Models (LLMs), which we define as Small Language Models (SLMs), on which previous CoT improvements cannot be fully observed \cite{shridhar-etal-2023-distilling,ho-etal-2023-large}. Therefore, several works are emerging at this gap, aiming to transfer LLM reasoning properties to SLMs. Pioneering proposals in this direction proposed teacher-student alignment methods through a series of approaches geared towards the distillation of the knowledge generated by the teacher for the fine-tuning of the student \cite{li-etal-2023-making,magister-etal-2023-teaching,shridhar-etal-2023-distilling}.
Later, \citet{yue2023mammoth} proposed specialized Instruction-tuning using Alpaca-like style demonstrations \cite{alpaca} specialized for mathematical tasks, while \citet{luo2023wizardmath,xu2023wizardlm} proposed supervised fine-tuning reinforced with rewarding algorithms.

\subsection{Reinforcement Learning (RL)}
\label{sec:RL_RelWork}
A significant component that promotes the generative reasoning delivering CoT is provided by refinement via RL methods. Recent work that applies Proximal Policy Optimization (PPO) \cite{schulman2017proximal} for aligning human preferences \cite{ouyang2022training}. Several methods have been proposed to improve the efficiency of alignment \cite{azar2023general}, including Direct Preference Optimization (DPO) \cite{rafailov2023direct}. 

In this work, we adopt RL to refine performance over conventional SFT. For mathematical problem solving, \citet{uesato2022solving} trained an outcome- or process-based reward model to perform re-ranking \cite{Cobbe2021TrainingVT}, achieving better performance than SFT and majority voting \cite{wang2023selfconsistency}.
\cite{luong2024reft} adopted reinforcement learning as an extension of traditional supervised tuning.
We adopt DPO and automate the reward process in a teacher-student context. We focus on the transfer of CoT-style, step-wise reasoning and propose a refinement technique applied to models downstream of the instruction-tuning phase.

\subsection{Self-refined Instruction-tuning}

Complementing and enhancing foundational approaches \cite{magister-etal-2023-teaching,uesato2022solving,li-etal-2023-symbolic,ho-etal-2023-large},  
several papers have been published simultaneously \citet{wang-etal-2023-democratizing,luo2023wizardmath,wang2023making,paul2024refiner,luong2024reft,ranaldi-freitas-2024-aligning} (Table \ref{tab:resume_rel_work} summarises the main features). These works prove the effect of supervised fine-tuning to transfer the ability to produce multi-step reasoned answers from larger to smaller models, as described in Section \ref{sec:RL_RelWork}. 
Our work goes beyond the state-of-the-art by:
\begin{itemize}
    \item proposing a method for aligning CoT abilities by introducing Instruction-tuning via Demonstrations produced by answers generated by different LLMs, decentralizing the unique teacher model (in many cases GPT-3.5,4).
    \item analyzing the alignment performance between in-family and out-family models on different tasks related to commonsense and math reasoning, identifying crucial alignment factors that arise between teachers and students.
    \item investigating the impact of teacher-student alignment by adapting and promoting DPO \cite{rafailov2023direct} as a cornerstone method for eliminating performance gaps. 
\end{itemize}

\section{Conclusion}
This paper proposes a novel approach for aligning multi-step CoT reasoning between teacher Large Language Models (LLMs) and student Smaller LMs (SLMs). In particular, our Self-refine Instruction-tuning is framed as an instruction tuning via Chain-of-Thought Demonstrations method based on explanations delivered by LLMs prompted by the CoT mechanism, which is then reinforced via the Self-refine phase that uses Direct Preference Optimization. We also contrast the impact of in-family and out-family alignment across teacher and student models. 
The results highlight the impact of teacher-student Instruction-tuning interventions as a mechanism to improve the multi-wise reasoning properties of smaller language models and promote the self-refinement abilities of instructed models to complete the alignment. 

\newpage

\section*{Limitations}
In this paper, we analyzed the impact of Answers delivered by Large Language Models using them as Demonstrations to reinforce the abilities of Small Language Models. Although we proposed an extensive study, there are several limitations:
\begin{itemize}
    \item only English-language prompting methods and tasks are considered. The understanding of these methods across different languages still needs to be established. 
    \item dependence on Large Language Models, where the supporting training sets are not always fully known. Although the characteristics of the corpora are reported in the system reports. Consequently, contextualising the differences in pre-training data between models is not fully possible, where the analysis is constrained to observing the outputs in natural language.
\end{itemize}
In conclusion, learning from and with Demonstrations carries some specific risks associated with automation. Although a model may generalize its predictions using a seemingly consistent series of natural language steps, even if the prediction is ultimately correct, there is no guarantee that the predicted output comes from a process represented by the generalization. A end-user might be overconfident in the model based on the CoT mechanism. 

\section*{Ethical Statement}
Although this research enhances the reasoning abilities of Smaller Language Models, they still need to be made sufficiently robust to be applied within more critical domains. Further safety and out-of-distribution generalisation mechanisms needs to be developed in tandem with the application of the methods described in this paper, in order to establish the robustness of the described mechanisms.

\bibliography{anthology,custom}

\appendix

\begin{table*}

\section{Instruction-tuning}
\label{sec:Appendix_Instruction-tuning}
The Instruction-tuning proposed in our contribution follows the pipeline proposed in \cite{ranaldi-freitas-2024-aligning} to achieving teacher-student alignment comprises two steps: annotation and knowledge transfer. In the annotation phase, Large Language Models (teachers) are prompted with questions (see Table \ref{tab:es_prompt_CoT}). The answers are collated and form the Demonstrations (see Table \ref{tab:esempio_input}). They then move on to the Instruction-tuning phase, conducted using what was proposed in \cite{alpaca}. In particular, the Demonstrations are constructed with triples formed by the instruction (a pattern to guide the generation related to the task), the input, which is the question related to the mathematical problem or the desired question, and the output the prompted LLM generated. Note that instruction and input can oftentimes be concatenated, but this depends on the basic configurations of the patterns and the type of task to be solved.
The instruction-tuning process, a specialization of task-oriented fine-tuning, is similar to the latter and can be described in Section \ref{sec:phase1}.

\section{Self-refine Instruction-tuning}
\label{sec:APP_DPO}
In order to refine Small Language Models (students) instructed via Demonstrations delivered by Large Language Models (teachers) we propose the Self-refine phase (introduced in Section \ref{sec:phase2}). In particular, this is based on a variant of the DPO optimization algorithm \cite{rafailov2023direct}.

Starting from the Demonstrations defined as $\mathcal{D} = (i_i, q_i, a_i)$ where $i \in \mathcal{D}$ (note that $a_i$ are generated using CoT prompt as showed in Appendix \ref{app:prompting}), we prompt the students using the input $x_i = i_i + q$ ( and $\hat{x}_{i} = x_i +$ \texttt{"Let's think step by step"}) $\forall i \in \mathcal{D}$.

Hence, for each element in Demonstrations, we collect the \textbf{Answers} ($y_i = \pi_{\text{inst}}(x_i)$) that are the answers generated by the student given the input $x_i$, and the \textbf{CoT-Answers} ($\hat{y}_{CoT} = \pi_{\text{inst}}(\hat{x}_{i})$) are the answers that deliver CoT generated by the student elicited via CoT mechanism $\hat{x}_{i}$. 

Hence, we introduce:
\begin{itemize}
    \item \textbf{Oracle or Target} $t_i$ that is the target answer given the input $x_i$.
    \item \textbf{Demonstration Answer} $\hat{a}_i$ and $a_i$: that are target answer given the input $x_i$ or $\hat{x}_{i}$.
    \item \textbf{Answer} $y_i = \pi_{\text{inst}}(x)$: is the answer generated by the student given the input $x$ (without CoT prompt).
    \item \textbf{CoT Answer} $y_{CoT} = \pi_{\text{inst}}(x_{CoT})$: is the answer that delivers CoT generated by the student elicited via CoT mechanism $x_{CoT}$.
\end{itemize}

In the following lines, we formalize the structuring of  \textbf{\texttt{DPO$_{CoT}$}}, \textbf{\texttt{DPO$_{answer}$}} and other configurations.

\paragraph{\textbf{\texttt{DPO$_{CoT}$}}} We propose \textbf{\texttt{DPO$_{CoT}$}} where the answers that deliver correct CoT are referred to as the preferred response, while the others are the answers without CoT defined as:

\begin{equation}
\begin{aligned}
\mathcal{L}_{\texttt{DPO}_{CoT}}(\pi_{\theta}; \pi_{\text{inst}}) = -\mathbb{E}_{(x_{CoT}, y_{w}, y_{l}) \sim D} 
\left[ \log \sigma (M(x_{CoT}, y_{w}, y_{l})) \right]
\end{aligned}
\label{eq:DPO_CoT}
\end{equation}

Where $\mathcal{L}_{\texttt{DPO}_{CoT}}(\pi_{theta}; \pi_{text{inst}})$ the same $\mathcal{L}_\texttt{DPO}$ introduced in Section \ref{sec:phase2} but in particular to elicit preferred generations the $y_{w}$ and $y_{l}$ components are defined as follows, $\forall i \in \mathcal{D}$ :

\begin{equation}
y_{w} = \begin{cases} 
\hat{y}_{i} & \text{if } t_i \in \hat{y}_{i} \\
\hat{a}_i & \\ 
\end{cases}
\end{equation}

while the discouraged answers are $y_{l}$ that are $y_i$ $\forall i \in \mathcal{D}$.

\paragraph{\textbf{\texttt{DPO$_{answer}$}}} In contrast, we propose \textbf{\texttt{DPO$_{answer}$}} and where the answers without CoT are referred to as the preferred.

\begin{equation}
\begin{aligned}
\mathcal{L}_{\texttt{DPO}_{answer}}(\pi_{\theta}; \pi_{\text{inst}}) = -\mathbb{E}_{(x, y_{p}, y_{CoT}) \sim D} 
\left[ \log \sigma (M(x, y_{p}, y_{CoT})) \right]
\end{aligned}
\label{eq:no_CoT}
\end{equation}

However, since our contribution is focused on CoT in the main work, we only consider \textbf{\texttt{DPO$_{CoT}$}}. In the Table \ref{fig:performances_increasing}, we have reported \textbf{\texttt{DPO$_{answer}$}} results.

\end{table*}

\begin{table*}[h]

\section{Experimental Details}
    \subsection{Data Splitting}
    \label{sec:experimental_details_splitting}
    In order to observe the impact of the Demonstrations, we produced a series of experiments by systematically decreasing the Instruction-tuning data.
    In particular, we chose three sub-sets with 75\%, 50\%, and 25\% from the total number of demonstrations.
    In detail, the Self-refine Instruction phases on the number of equal Demonstrations are performed by taking about 3000 examples in splitting 100\%, 2250 in splitting 50\%, 1500 in splitting 50\%, and 750 in splitting 25\%. We chose the value 3000 because it has the smallest CoT Demonstrations available. For the total Demonstrations, we selected random samples. Using these splitting, we performed the evaluations incrementally as the demonstrations used to do Instruction-tuning, to do Self-refine, and to do Self-refine Instruction-tuning.

        \subsection{Parameters}
    \label{sec:parameters}
    The annotation phase that the Teachers performed was done on the training set. The evaluation phase of both the basic models and the Students and the Teachers was done on the test splitting. The evaluation, described in Section \ref{sec:Evaluation}, was done with question probing and string matching of the generated answers. More specifically: 
    \paragraph{Teachers}
    We performed the annotation phase for each benchmark by delivering to \texttt{GPT-3.5-turbo}, \texttt{Mixtral7x8} and \texttt{Llama-2-70-chat}  the prompts structured as shown in Table \ref{tab:es_prompt} and Table \ref{tab:es_prompt_CoT} (customized for each benchmark). We set the temperatures to 0.7 for \texttt{GPT-3.5-turbo} and 0.1 for \texttt{Llama-2-70-chat} as recommended in technical reports. Moreover, we kept all the other parameters as default. All parameters are shown in our code \url{}.
    \paragraph{Baseline \& Students}
    We evaluated the performance of the Small Language Models (\texttt{Llama-2-7-chat}, \texttt{Llama-2-13-chat}, \texttt{Mistral-7b}) by prompting them with the same format used for the Teachers. For both the baselines and the instructed models, we set the temperature to 0.1 and kept all the other parameters as default. The evaluation pipelines and generation parameters are available in our code.

\end{table*}

\begin{table*}[t]
\section{Prompting Approaches}
\label{app:prompting}
\noindent
{\setlength{\fboxsep}{-1.4pt} 
\colorbox{lightgray}{ 
\begin{tabular}[t]{|p{0.45\textwidth}|}
    \hline
    \textit{Prompt for task:} OBQA, CSQA, PIQA, SIQA  \\
    \hline
    \texttt{\textbf{Choose the answer to the question only from options A, B, C, [...].}} \\
     \texttt{\textbf{Question:} <Question>}\\
\texttt{\textbf{Choices:}}\\
\texttt{A) <Option1>}\\
\texttt{B) <Option2>}\\
\texttt{C) <Option3>}\\
\texttt{....}\\
\texttt{\textbf{Answer:}} \\
    \hline
    \end{tabular}
    }
    }
\hfill
{\setlength{\fboxsep}{-1.4pt} 
\colorbox{lightgray}{
\begin{tabular}[t]{|p{0.45\textwidth}|}
    \hline
    \textit{Prompt for task:} GSM8k, MultiArith  \\
    \hline
    \texttt{\textbf{Answer the following mathematical question with numerical solution.}} \\
     \texttt{\textbf{Question:} <Question>}\\
\texttt{\textbf{Answer:}} \\
    \hline
\end{tabular}
    } 
    }
\caption{Example of input-prompt for multiple-choices (left) and mathematical (right) question-answering benchmarks.}
\label{tab:es_prompt}

\noindent
{\setlength{\fboxsep}{-1.4pt} 
\colorbox{lightgray}{ 
\begin{tabular}[t]{|p{0.45\textwidth}|}
    \hline
    \textit{Prompt for task:} OBQA, CSQA, PIQA, SIQA  \\
    \hline
    \texttt{\textbf{Choose the answer to the question only from options A, B, C, [...].}} \\
     \texttt{\textbf{Question:} <Question>}\\
\texttt{\textbf{Choices:}}\\
\texttt{A) <Option1>}\\
\texttt{B) <Option2>}\\
\texttt{C) <Option3>}\\
\texttt{....}\\
\texttt{\textbf{Answer: \uline{Let's think step by step}}} \\
    \hline
    \end{tabular}
    } % End colorbox
    }
\hfill
{\setlength{\fboxsep}{-1.4pt} 
\colorbox{lightgray}{ % Start colorbox
\begin{tabular}[t]{|p{0.45\textwidth}|}
    \hline
    \textit{Prompt for task:} GSM8k, MultiArith  \\
    \hline
    \texttt{\textbf{Answer the following mathematical question with numerical solution.}} \\
     \texttt{\textbf{Question:} <Question>}\\
\texttt{\textbf{Answer: \uline{Let's think step by step}}} \\
    \hline
\end{tabular}
    } % End colorbox
    }
\caption{Example \textbf{Zero-shot CoT} of input-prompt for multiple-choices (left) and mathematical (right) question-answering benchmarks (approach used in this work).}
\label{tab:es_prompt_CoT}

\end{table*}

\begin{table*}
    
\section{Models}
\label{app:B}

\centering 
\begin{tabular}{l|l}
\textbf{Model} & \textbf{Version}  \\ 

\hline
\hline

\hline
Llama-2-7-chat   &  meta-llama/Llama-2-7b \\
Llama-2-13-chat   &  meta-llama/Llama-2-13b \\
Llama-2-70-chat   &  meta-llama/Llama-2-70b \\
Mistral-7   & mistralai/Mistral-7B-Instruct-v0.1  \\
Mixtral7x8   & mistralai/Mixtral-8x7B-v0.1  \\
\hline
\end{tabular}

\caption{List and specific versions of the models proposed in this work, which can be found on \url{huggingface.co}. For each model we used all the default configurations proposed in the repositories.}
\label{tab:versions_models_HF}

\end{table*}

\begin{table*}[]
\section{Accuracy of LLMs on different Benchhmark}
\centering
\begin{tabular}{l|cc|cc|cc}
\multicolumn{1}{l}{\textbf{Benchmarks}} & \multicolumn{2}{c|}{\textbf{Llama-2-70}} & \multicolumn{2}{c}{\textbf{GPT-3.5}} & \multicolumn{2}{c}{\textbf{Mixtral7x8}} \\

 & \textbf{Baseline} & \textbf{CoT} & \textbf{Baseline} & \textbf{CoT} & \textbf{Baseline} & \textbf{CoT} \\
\midrule
\textbf{Training} &  & & & & & \\
\midrule
OpenBook QA & 65.6{\tiny $\pm .3$} & 71.3{\tiny $\pm .1$} & 66.2{\tiny $\pm .2$} & 75.4{\tiny $\pm .4$} & 77.9{\tiny $\pm .3$} & \textbf{81.2}{\tiny $\pm .1$} \\
CommonSesnse QA & 74.2{\tiny $\pm .1$} & 79.6{\tiny $\pm .3$} & 79.3{\tiny $\pm .4$} & \textbf{84.8}{\tiny $\pm .1$} & 78.2{\tiny $\pm .2$} & 82.3{\tiny $\pm .3$} \\
\hline
Social Interaction QA & 65.4{\tiny $\pm .2$} & 67.5{\tiny $\pm .3$} & 67.6{\tiny $\pm .5$} & \textbf{70.3}{\tiny $\pm .4$} & 65.5{\tiny $\pm .2$} & 68.2{\tiny $\pm .3$} \\
Physical Interaction QA & 82.6{\tiny $\pm .2$} & \textbf{85.8}{\tiny $\pm .2$}{\tiny $\pm .3$} & 83.5{\tiny $\pm .3$} & 85.3{\tiny $\pm .1$} & 80.2{\tiny $\pm .3$} & 84.1{\tiny $\pm .3$} \\
\hline
GSM8K & 74.6{\tiny $\pm .1$} & 77.2{\tiny $\pm .2$} & 83.2{\tiny $\pm .2$} & \textbf{86.5}{\tiny $\pm .1$} & 65.6{\tiny $\pm .4$} & 67.9{\tiny $\pm .2$} \\
MultiArith & 88.6{\tiny $\pm .4$} & 90.8{\tiny $\pm .3$} & 94.9{\tiny $\pm .4$} & \textbf{96.7}{\tiny $\pm .1$} & 89.3{\tiny $\pm .1$} & 91.5{\tiny $\pm .4$} \\
\midrule
\textbf{Testing} &  & & & & & \\
\midrule
OpenBook QA & 65.9{\tiny $\pm .2$} & 70.8{\tiny $\pm .1$} & 67.8{\tiny $\pm .1$} & 74.6{\tiny $\pm .4$} & 78.4{\tiny $\pm .3$} & \textbf{84.6}{\tiny $\pm .2$} \\
CommonSesnse QA & 73.4{\tiny $\pm .2$} & 81.8{\tiny $\pm .3$} & 80.2{\tiny $\pm .2$} & \textbf{83.7}{\tiny $\pm .1$} & 77.6{\tiny $\pm .3$} & 81.5{\tiny $\pm .1$} \\
\hline
Social Interaction QA & 64.2{\tiny $\pm .2$} & 66.9{\tiny $\pm .4$} & 66.9 & \textbf{71.3}{\tiny $\pm .3$} & 64.3{\tiny $\pm .3$} & 65.4{\tiny $\pm .2$} \\
Physical Interaction QA & 82.6{\tiny $\pm .3$} & 85.6{\tiny $\pm .5$}  & 84.3{\tiny $\pm .2$} & \textbf{85.8}{\tiny $\pm .5$} & 79.9{\tiny $\pm .3$} & 83.5{\tiny $\pm .1$} \\
\hline
GSM8K & 75.2{\tiny $\pm .5$} & 77.8{\tiny $\pm .5$} & 82.8{\tiny $\pm .2$} & \textbf{84.6}{\tiny $\pm .4$} & 63.4{\tiny $\pm .3$} & 62.8{\tiny $\pm .5$} \\
MultiArith & 89.2{\tiny $\pm .1$} & 92.3{\tiny $\pm .2$} & 95.6{\tiny $\pm .2$} & \textbf{97.4}{\tiny $\pm 3$} & 88.9{\tiny $\pm .1$} & 90.2{\tiny $\pm .3$} \\
\bottomrule
\end{tabular}
\caption{Accuracy (\%) of Llama-2-70, GPT-3.5 and Mixtral7x8 (teachers) on training and testing data with CoT prompt (CoT) and with the standard prompt (Baseline).}
\label{tab:evaluations_teachers}

\end{table*}

\begin{table*}[t]
\small
\centering
\begin{tabular}{|p{0.9\linewidth}|}
\multicolumn{1}{l}{\texttt{\textbf{Instruction}}} \\
\hline
\texttt{Choose the answer to the question only from options A, B, C, D.} \\
\hline
\end{tabular}

\vspace{0.5em} 

\begin{tabular}{|p{0.9\linewidth}|}
\multicolumn{1}{l}{\texttt{\textbf{Input}}} \\
\hline
\texttt{\textit{Question:} Which form can be made into a triangle?} \\ 
\texttt{Choices:} \\ 
\texttt{A) ice} \\ 
\texttt{B) steam} \\ 
\texttt{C) water} \\ 
\texttt{D) salt water} \\
\texttt{Answer: \textbf{Let's think step-by-step}} \\
\hline
\end{tabular}

\vspace{0.5em}

\begin{tabular}{|p{0.9\linewidth}|}
\multicolumn{1}{l}{\texttt{\textbf{Output}} (that is the \texttt{Answer} produced by teacher model)} \\
\hline
\texttt{\textit{Answer:}} \\
\texttt{A) ice: Ice is a solid form of water. In solid form, substances can be molded or shaped into various forms, including triangles. If you had a block of ice, you could carve or mold it into a triangular shape.} \\
\texttt{B) steam: Steam is the gaseous form of water. Being a gas, it doesn't have a fixed shape. You can't shape steam into a triangle without containing it within some form of solid boundary, and even then, the steam itself isn't triangular - the container is.} \\
\texttt{C) water: Water is a liquid that takes the shape of its container. You can pour water into a triangular container, and while the volume inside would be triangular, the water itself isn't inherently triangular.} \\

\texttt{D) salt water: This is essentially the same as option C, but with salt dissolved in it. Being a liquid, it takes the shape of its container.} \\

\texttt{Given these considerations, the only substance among the options that can be shaped into a triangle by itself is \textbf{A) Ice}.} \\
\hline
\end{tabular}

\caption{Example of Demonstration produced by the GPT-3.5 answer on an Open Book Question Answering benchmark instance. The structure is composed by: \texttt{\textbf{Instruction}}, \texttt{\textbf{Input}} and \texttt{\textbf{Output}}.
}
\label{tab:esempio_input}
\end{table*}

\begin{table*}[]

\section{Description of proposed Benchmark}

\centering 
\small
\begin{tabular}{l|c}

\textbf{Dataset} & \textbf{Example}  \\ 

\hline
\hline

Open Book Question Answering & \textit{When birds migrate south for the winter, they do it because}  \\
(OBQA) \cite{mihaylov2018suit} & \textbf{A) they are genetically called to.} B) their children ask them to. 
  \\
 &  C) it is important to their happiness. D) they decide to each.	
  \\
\hline

Common Sense Question Answering & \textit{Aside from water and nourishment what does your dog need?}	
 \\
(CSQA) \cite{talmor-etal-2019-commonsenseqa} &  A) bone. B) charm. C) petted.  \\
 &\textbf{ D) lots of attention.} E) walked. \\

\hline

Physical Interaction Question Answering  & \textit{How do you attach toilet paper to a glass jar?} 	\textbf{A) Press a piece of double-sided}
 \\ 

(PIQA) \cite{bisk2019piqa} &  \textbf{tape to the glass jar and then press the toilet paper onto the tape.} \\
 & B) Spread mayonnaise all over the jar with your palms and then roll the jar in toilet paper. \\

\hline

Social Interaction Question Answering & \textit{Taylor gave help to a friend who was having trouble keeping up with their bills.}   \\
(SIQA) \cite{sap-etal-2019-social} &  \textit{What will their friend want to do next?} A) Help the friend find a higher  \\
  & paying job. \textbf{B) Thank Taylor for the generosity.} 
C) pay some of their late employees. \\
\hline

 & Tina makes \$18.00 an hour. If she works more than 8 hours per shift,   \\
(GSM8K) \cite{Cobbe2021TrainingVT} &  she is eligible for overtime, which is paid by your  wage + 1/2 your hourly \\ & 
hourly wage. If she works 10 hours every day for 5 days,  \\
 &  how much money does she make?   \\
\hline

 & Chloe was playing a video game where she scores 9 points for each    \\
(MultiArith) \cite{roy-roth-2015-solving} &  treasure she finds. If she found 6 treasures on the  \\ 
& first level and 3 on the second, \\
 &  what would her score be?   \\
\hline

\end{tabular}

\caption{Examples of the benchmarks used in this paper. }
\label{tab:examples_benchmarks}

\vskip 1cm

\centering 
\begin{tabular}{lcccc|cc}
\toprule[1.2pt]
 & \textbf{OBQA} & \textbf{CSQA} & \textbf{PIQA} & \textbf{SIQA} & \textbf{GSM8K} & \textbf{MultiArith} \\
\midrule[1pt]
classes & 4 & 5 & 2 & 3 & - & - \\
\midrule
\textbf{Training} &  &  &  & & & \\
\# examples for & 1000 & 800 & 2000 & 1330 & 4000 & 420 \\
each class &  &  &  & &  & \\
\midrule
\textbf{Test} &  &  &  & & & \\
\# examples for & 125$^*$ & 235$^*$  & 924$^*$ & 640$^*$ & 1318 & 180 \\
each class & ($\pm$ 8) & ($\pm$ 11) & ($\pm$ 18) & ($\pm$ 19) &  & \\
\midrule
\bottomrule[1.5pt]
\end{tabular}
\caption{Characteristics Training and Test set of benchmarks proposed in Section \ref{sec:data}. The * indicates that the number of examples are not perfect balanced, but the difference from the average is marginal. GMS8K e MultiArith are not closed-ended question answering; they only have a question and a numerical solution. }
\label{tab:dataset_summary}

\vskip 1cm

\centering 
\begin{tabular}{l|l}
\textbf{Name} & \textbf{Repository}  \\ 

\hline
\hline

\hline
CommonSenseQA \cite{talmor-etal-2019-commonsenseqa}  & \url{huggingface.co/datasets/commonsense_qa} \\
OpenBookQA \cite{mihaylov2018suit} &  \url{huggingface.co/datasets/openbookqa} \\
StrategyQA \cite{} & \url{huggingface.co/datasets/voidful/StrategyQA} \\
\hline
PIQA  \cite{bisk2019piqa} &  \url{huggingface.co/datasets/piqa} \\
SIQA \cite{sap-etal-2019-social} & \url{huggingface.co/datasets/social_i_qa}  \\
\hline
GSM8K \cite{Cobbe2021TrainingVT} & \url{huggingface.co/datasets/gsm8k} \\
MultiArith \cite{roy-roth-2015-solving} & \url{huggingface.co/datasets/ChilleD/MultiArith} \\
\hline
\end{tabular}

\caption{In this table, we list the versions of the benchmark proposed in this work, which can be found on huggingface.co.}
\label{tab:versions_data_HF}

\end{table*}

\begin{table*}[t]
\small
\centering
\begin{tabular}{l|c|ccccccc}
\toprule
\textbf{Trained on} & \textbf{Teacher} & \multicolumn{6}{c}{\textbf{Evaluated on}} \\
\cmidrule{3-8}
& & \textbf{OBQA} & \textbf{CSQA} & \textbf{PIQA} & \textbf{SIQA} & \textbf{GMS8K} & \textbf{MultiArith} \\
\midrule
{Baseline} & - & 55.4{\tiny $\pm .2$} & 63.4{\tiny $\pm .3$} & 66.4{\tiny $\pm .2$} & 48.3{\tiny $\pm .2$} & 65.6{\tiny $\pm .4$} & 63.4{\tiny $\pm .2$} \\
{Baseline CoT} & - & 54.2{\tiny $\pm .2$} & 62.8{\tiny $\pm .4$} & 71.2{\tiny $\pm .3$} & 46.9{\tiny $\pm .5$} & 70.5{\tiny $\pm .1$} & 62.8{\tiny $\pm .2$} \\
\hline

\multirow{3}{*}{\textbf{OBQA}} &  \texttt{Instruction-tuning} & \cellcolor{inndomain}68.5{\tiny $\pm .4$} & \cellcolor{indomain}67.5{\tiny $\pm .3$} & \cellcolor{indomain}69.4{\tiny $\pm .1$} & \cellcolor{indomain}60.1{\tiny $\pm .2$} & \cellcolor{outdomain}62.3{\tiny $\pm .4$} & \cellcolor{outdomain}61.5{\tiny $\pm .5$} \\
& \texttt{+ Self-refine} & \cellcolor{inndomain}\textbf{71.2}{\tiny $\pm .4$} & \cellcolor{indomain}74.1{\tiny $\pm .2$} & 
\cellcolor{indomain}76.2{\tiny $\pm .3$} & \cellcolor{indomain}63.4{\tiny $\pm .3$} & \cellcolor{outdomain}69.9{\tiny $\pm .4$} & \cellcolor{outdomain}70.7{\tiny $\pm .2$} \\
& \texttt{Cross Self-refine} & - & 79.2{\tiny $\pm .1$} & 
79.5{\tiny $\pm .2$} & 65.6{\tiny $\pm .3$} & 75.2{\tiny $\pm .4$} & 84.3{\tiny $\pm .5$} \\

\hline

\multirow{3}{*}{\textbf{CSQA}}
 & \texttt{Instruction-tuning} & \cellcolor{indomain}58.4{\tiny $\pm .4$} & \cellcolor{inndomain}77.5{\tiny $\pm .2$} & \cellcolor{indomain}66.4{\tiny $\pm .2$} & \cellcolor{indomain}61.8{\tiny $\pm .3$} & \cellcolor{outdomain}62.4{\tiny $\pm .4$} & \cellcolor{outdomain}60.2{\tiny $\pm .2$} \\
 &\texttt{+ Self-refine} & \cellcolor{indomain}69.5{\tiny $\pm .5$} & \cellcolor{inndomain}\textbf{81.4}{\tiny $\pm .2$} & \cellcolor{indomain}74.2{\tiny $\pm .5$} & \cellcolor{indomain}67.9{\tiny $\pm .1$} & \cellcolor{outdomain}62.1{\tiny $\pm .3$} & \cellcolor{outdomain}61.4{\tiny $\pm .4$} \\
 &\texttt{Cross Self-refine} & 70.2{\tiny $\pm .4$} & - & 79.5{\tiny $\pm .3$} & 65.2{\tiny $\pm .1$} & 73.3{\tiny $\pm .3$} & 75.3{\tiny $\pm .5$} \\
\hline

\multirow{3}{*}{\textbf{PIQA}} & \texttt{Instruction-tuning} & \cellcolor{indomain}57.8{\tiny $\pm .2$} & \cellcolor{indomain} 65.2{\tiny $\pm .3$} & \cellcolor{inndomain} 81.9{\tiny $\pm .4$} & \cellcolor{indomain} 58.5{\tiny $\pm .4$} & \cellcolor{outdomain} 59.2{\tiny $\pm .4$} & \cellcolor{outdomain} 60.3{\tiny $\pm .3$} \\
 & \texttt{+ Self-refine} & \cellcolor{indomain}69.6{\tiny $\pm .2$} & \cellcolor{indomain} 68.2{\tiny $\pm .4$} & \cellcolor{inndomain}\textbf{85.1}{\tiny $\pm .5$} & \cellcolor{indomain} 64.3{\tiny $\pm .1$} & \cellcolor{outdomain} 69.3{\tiny $\pm .2$} & \cellcolor{outdomain} 68.1{\tiny $\pm .3$} \\
 & \texttt{Cross Self-refine} & 69.9{\tiny $\pm .1$} &  71.3{\tiny $\pm .1$} &  - &  65.3{\tiny $\pm .1$} &  69.6{\tiny $\pm .4$} &  69.2{\tiny $\pm .2$} \\
\hline

\multirow{3}{*}{\textbf{SIQA}} & \texttt{Instruction-tuning} & \cellcolor{indomain}59.6{\tiny $\pm .1$} & \cellcolor{indomain} 63.9{\tiny $\pm .4$} & \cellcolor{indomain} 67.1{\tiny $\pm .2$} & \cellcolor{inndomain} 64.5{\tiny $\pm .3$} & \cellcolor{outdomain} 60.3{\tiny $\pm .4$} & \cellcolor{outdomain} 61.3{\tiny $\pm .2$} \\

& \texttt{+ Self-refine} & \cellcolor{indomain} 69.2{\tiny $\pm .2$} & \cellcolor{indomain} 69.4{\tiny $\pm .1$} & \cellcolor{indomain} 79.2{\tiny $\pm .4$} & \cellcolor{inndomain}\textbf{66.7}{\tiny $\pm .3$} & \cellcolor{outdomain}62.4{\tiny $\pm .4$} & \cellcolor{outdomain}61.8{\tiny $\pm .2$} \\
& \texttt{Cross Self-refine} &  71.2{\tiny $\pm .2$} &  69.2{\tiny $\pm .1$} &  80.4{\tiny $\pm .2$} & - & 66.5{\tiny $\pm .1$} & 66.7{\tiny $\pm .2$} \\
\hline

\multirow{3}{*}{\textbf{GSM8K}} & \texttt{Instruction-tuning} & \cellcolor{outdomain}54.3{\tiny $\pm .2$} & \cellcolor{outdomain}55.8{\tiny $\pm .3$} & \cellcolor{outdomain}64.3{\tiny $\pm .4$} & \cellcolor{outdomain}53.2{\tiny $\pm .3$} & \cellcolor{inndomain}72.3{\tiny $\pm .3$} & \cellcolor{indomain}71.6{\tiny $\pm .2$} \\
& \texttt{+ Self-refine} & \cellcolor{outdomain}59.3{\tiny $\pm .4$} & \cellcolor{outdomain}62.2{\tiny $\pm .2$} & \cellcolor{outdomain}63.5{\tiny $\pm .3$} & \cellcolor{outdomain}53.5{\tiny $\pm .5$} & \cellcolor{inndomain}\textbf{77.2}{\tiny $\pm .4$} & \cellcolor{indomain}75.2{\tiny $\pm .3$} \\
& \texttt{Cross Self-refine} & 65.7{\tiny $\pm .1$} & 65.2{\tiny $\pm .5$} & 78.1{\tiny $\pm .3$} & 61.6{\tiny $\pm .4$} & - & 76.2{\tiny $\pm .2$} \\
\hline

\multirow{3}{*}{\textbf{MultiArith}} & \texttt{Instruction-tuning} & \cellcolor{outdomain}54.7{\tiny $\pm .2$} & \cellcolor{outdomain}56.6{\tiny $\pm .3$} & \cellcolor{outdomain}54.5{\tiny $\pm .3$} & \cellcolor{outdomain}52.4{\tiny $\pm .3$} & \cellcolor{indomain}70.2{\tiny $\pm .1$} & \cellcolor{inndomain}75.8{\tiny $\pm .2$} \\
& \texttt{+ Self-refine} & \cellcolor{outdomain}60.3{\tiny $\pm .2$} & \cellcolor{outdomain}64.1{\tiny $\pm .4$} & \cellcolor{outdomain}59.4{\tiny $\pm .3$} & \cellcolor{outdomain}59.7{\tiny $\pm .1$} & \cellcolor{indomain}72.1{\tiny $\pm .4$} & \cellcolor{inndomain}\textbf{86.2}{\tiny $\pm .3$} \\
& \texttt{Cross Self-refine} & 66.2{\tiny $\pm .3$} & 62.4{\tiny $\pm .1$} & 63.2{\tiny $\pm .3$} & 61.5{\tiny $\pm .4$} & 73.9{\tiny $\pm .2$} & - \\

\bottomrule
\end{tabular}
\caption{Evaluation of Llama-2-13 Instruction-tuned (\texttt{Instruction-tuned}) and with completely Self-refine Instruction-tuning (\texttt{+ Self-refine Instruction-tuned}) on Demonstrations using different test sets. We evaluate in-domain (QA vs QA) and out-domain (QA vs math-word problem) benchmarks. "Baselines" are referred to the non-instructed model. Results colored in green indicate the in-domain benchmark, blue the out-domain benchmark, and orange the same benchmark on which the evaluation phase is performed. Moreover, we propose Self-refine Instruction-tuning in cross-setting scenarios where we optimize the model on the training set related to the evaluated task.}
\label{tab:generalization_results_llama_13}
\end{table*}

\begin{figure*}[t]
\centering
         \begin{minipage}{0.24\linewidth}
     \centering
     \includegraphics[width=\linewidth]{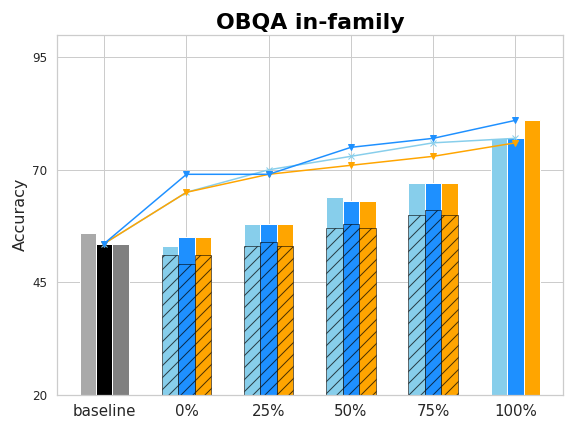}
   \end{minipage}
            \begin{minipage}{0.24\linewidth}
     \centering
     \includegraphics[width=\linewidth]{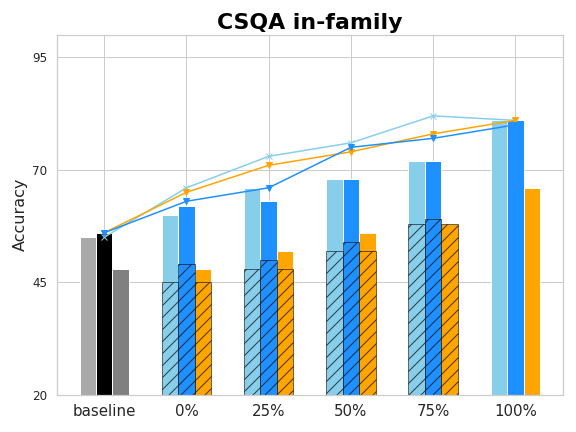}
   \end{minipage}
         \begin{minipage}{0.24\linewidth}
     \centering
     \includegraphics[width=\linewidth]{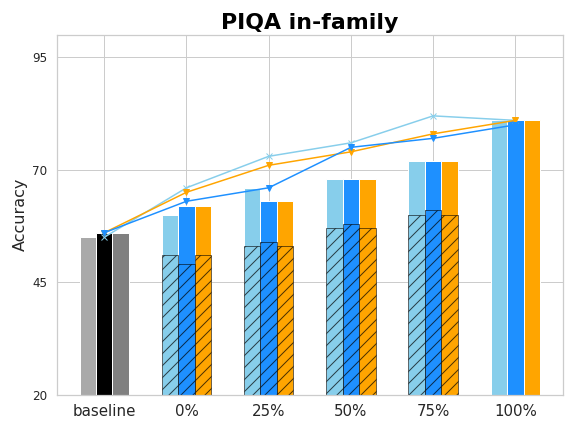}
   \end{minipage}
            \begin{minipage}{0.24\linewidth}
     \centering
     \includegraphics[width=\linewidth]{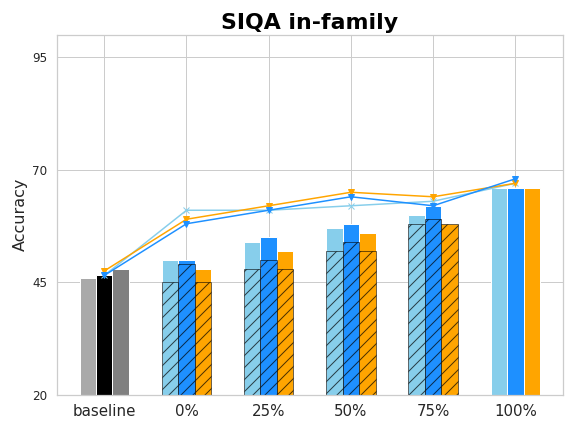}
   \end{minipage}

         \begin{minipage}{0.24\linewidth}
     \centering
     \includegraphics[width=\linewidth]{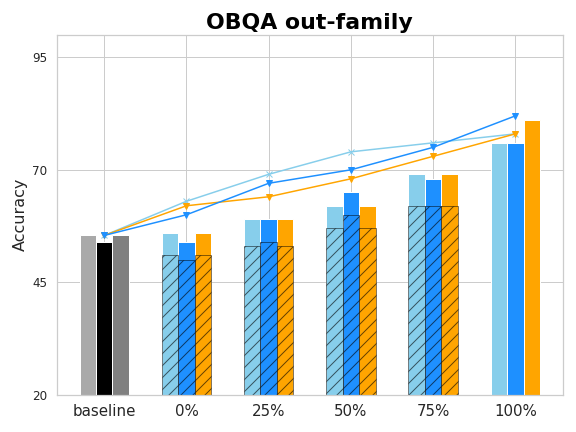}
   \end{minipage}
            \begin{minipage}{0.24\linewidth}
     \centering
     \includegraphics[width=\linewidth]{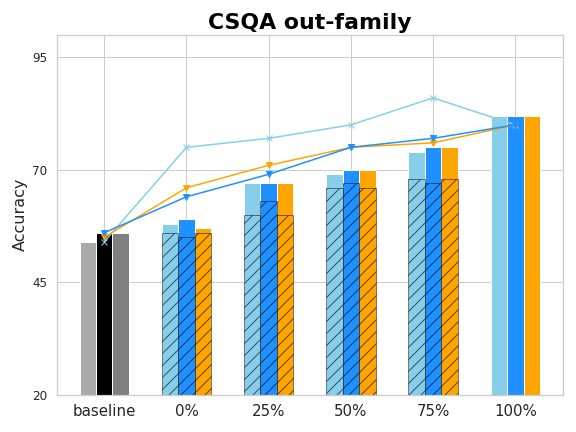}
   \end{minipage}
         \begin{minipage}{0.24\linewidth}
     \centering
     \includegraphics[width=\linewidth]{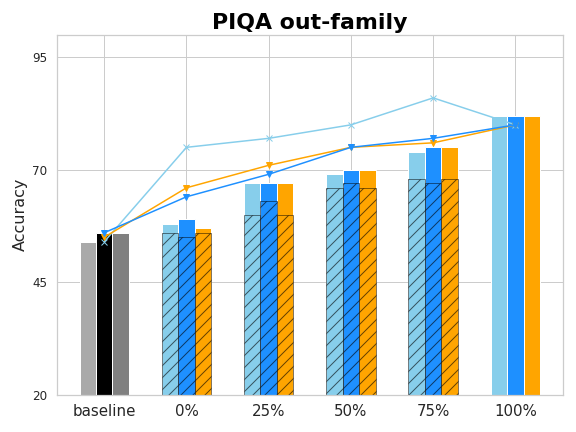}
   \end{minipage}
            \begin{minipage}{0.24\linewidth}
     \centering
     \includegraphics[width=\linewidth]{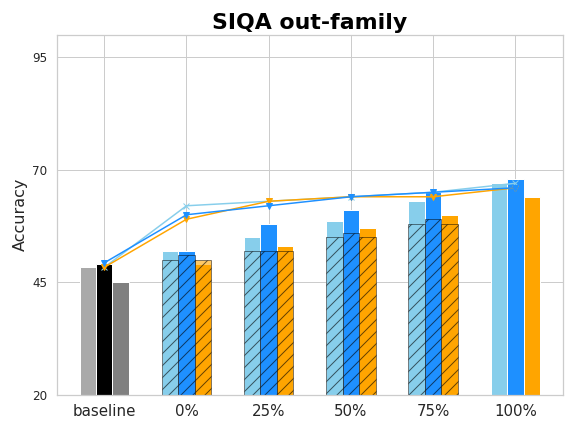}
   \end{minipage}

            \begin{minipage}{0.85\linewidth}
     \centering
     \includegraphics[width=\linewidth]{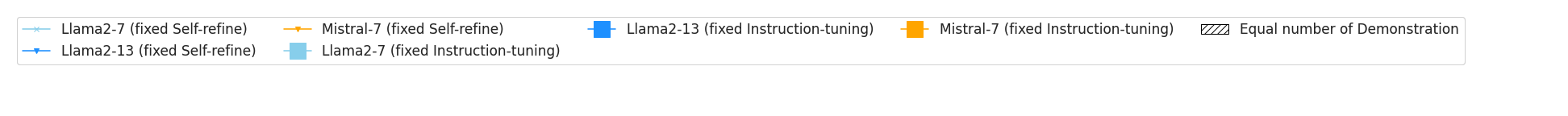}
   \end{minipage}

   \caption{Acciracies (\%) on the test set of benchmarks. The Self-refine Instruction-tuning performed on different splits (see Appendix \ref{sec:experimental_details_splitting} for major details). } 
   \label{fig:performances_increasing}

\end{figure*}

\begin{table*}[t]
\small
\centering
\begin{tabular}{l|c|ccccccc}
\toprule
\textbf{Trained on} & \textbf{Teacher} & \multicolumn{6}{c}{\textbf{Evaluated on}} \\
\cmidrule{3-8}
& & \textbf{OBQA} & \textbf{CSQA} & \textbf{PIQA} & \textbf{SIQA} & \textbf{GMS8K} & \textbf{MultiArith} \\
\midrule
{Baseline} & - & 62.7{\tiny $\pm .3$} & 69.2{\tiny $\pm .4$} & 67.3{\tiny $\pm .1$} & 55.3{\tiny $\pm .2$} & 54.2{\tiny $\pm .2$} & 88.4{\tiny $\pm .1$} \\
{Baseline CoT} & - & 60.4{\tiny $\pm .3$} & 68.7{\tiny $\pm .2$} & 66.1{\tiny $\pm .2$} & 54.8{\tiny $\pm .4$} & 55.6{\tiny $\pm .3$} & 87.3{\tiny $\pm .2$} \\
\hline
\multirow{3}{*}{\textbf{OBQA}} &  \texttt{Instruction-tuning} & \cellcolor{inndomain}78.3{\tiny $\pm .2$} & \cellcolor{indomain}65.4{\tiny $\pm .2$} & \cellcolor{indomain}67.2{\tiny $\pm .3$} & \cellcolor{indomain}59.2{\tiny $\pm .1$} & \cellcolor{outdomain}64.2{\tiny $\pm .2$} & \cellcolor{outdomain}62.1{\tiny $\pm .3$} \\
& \texttt{+ Self-refine} & \cellcolor{inndomain}\textbf{87.6}{\tiny $\pm .2$} & \cellcolor{indomain}73.1{\tiny $\pm .2$} & 
\cellcolor{indomain}76.1{\tiny $\pm .1$} & \cellcolor{indomain}63.3{\tiny $\pm .3$} & \cellcolor{outdomain}69.1{\tiny $\pm .4$} & \cellcolor{outdomain}70.7{\tiny $\pm .3$} \\
& \texttt{Cross Self-refine} & - & 79.4{\tiny $\pm .1$} & 
80.1{\tiny $\pm .2$} & 68.2{\tiny $\pm .4$} & 75.2{\tiny $\pm .4$} & 81.3{\tiny $\pm .1$} \\

\hline

\multirow{3}{*}{\textbf{CSQA}}
 & \texttt{Instruction-tuning} & \cellcolor{indomain}58.9{\tiny $\pm .1$} & \cellcolor{inndomain}73.1{\tiny $\pm .4$} & \cellcolor{indomain}65.8{\tiny $\pm .2$} & \cellcolor{indomain}62.1{\tiny $\pm .1$} & \cellcolor{outdomain}62.2{\tiny $\pm .3$} & \cellcolor{outdomain}60.2{\tiny $\pm .2$} \\
 &\texttt{+ Self-refine} & \cellcolor{indomain}69.5{\tiny $\pm .5$} & \cellcolor{inndomain}\textbf{81.3}{\tiny $\pm .1$} & \cellcolor{indomain}75.1{\tiny $\pm .1$} & \cellcolor{indomain}66.5{\tiny $\pm .2$} & \cellcolor{outdomain}61.1{\tiny $\pm .4$} & \cellcolor{outdomain}62.4{\tiny $\pm .1$} \\
 &\texttt{Cross Self-refine} & 69.2{\tiny $\pm .2$} & - & 79.3{\tiny $\pm .1$} & 65.2{\tiny $\pm .4$} & 72.8{\tiny $\pm .4$} & 74.4{\tiny $\pm .2$} \\
\hline

\multirow{3}{*}{\textbf{PIQA}} & \texttt{Instruction-tuning} & \cellcolor{indomain}58.6{\tiny $\pm .2$} & \cellcolor{indomain} 64.8{\tiny $\pm .2$} & \cellcolor{inndomain} 81.6{\tiny $\pm .2$} & \cellcolor{indomain} 59.2{\tiny $\pm .4$} & \cellcolor{outdomain} 60.2{\tiny $\pm .2$} & \cellcolor{outdomain} 60.3{\tiny $\pm .4$} \\
 & \texttt{+ Self-refine} & \cellcolor{indomain}68.2{\tiny $\pm .4$} & \cellcolor{indomain} 68.2{\tiny $\pm .5$} & \cellcolor{inndomain}\textbf{85.6}{\tiny $\pm .2$} & \cellcolor{indomain} 63.8{\tiny $\pm .2$} & \cellcolor{outdomain} 67.9{\tiny $\pm .2$} & \cellcolor{outdomain} 67.2{\tiny $\pm .4$} \\
 & \texttt{Cross Self-refine} & 69.2{\tiny $\pm .3$} &  71.9{\tiny $\pm .3$} &  - &  63.2{\tiny $\pm .1$} &  68.4{\tiny $\pm .5$} &  69.6{\tiny $\pm .1$} \\
\hline

\multirow{3}{*}{\textbf{SIQA}} & \texttt{Instruction-tuning} & \cellcolor{indomain}59.3{\tiny $\pm .2$} & \cellcolor{indomain} 66.8{\tiny $\pm .2$} & \cellcolor{indomain} 63.2{\tiny $\pm .4$} & \cellcolor{inndomain} 61.5{\tiny $\pm .2$} & \cellcolor{outdomain} 60.2{\tiny $\pm .1$} & \cellcolor{outdomain} 61.3{\tiny $\pm .3$} \\

& \texttt{+ Self-refine} & \cellcolor{indomain} 68.3{\tiny $\pm .3$} & \cellcolor{indomain} 68.5{\tiny $\pm .2$} & \cellcolor{indomain} 78.3{\tiny $\pm .3$} & \cellcolor{inndomain}\textbf{65.8}{\tiny $\pm .4$} & \cellcolor{outdomain}62.4{\tiny $\pm .5$} & \cellcolor{outdomain}61.3{\tiny $\pm .4$} \\
& \texttt{Cross Self-refine} &  71.3{\tiny $\pm .4$} &  69.2{\tiny $\pm .2$} &  78.1{\tiny $\pm .2$} & - & 65.6{\tiny $\pm .3$} & 68.3{\tiny $\pm .1$} \\
\hline

\multirow{3}{*}{\textbf{GSM8K}} & \texttt{Instruction-tuning} & \cellcolor{outdomain}52.4{\tiny $\pm .1$} & \cellcolor{outdomain}54.9{\tiny $\pm .5$} & \cellcolor{outdomain}58.7{\tiny $\pm .1$} & \cellcolor{outdomain}51.8{\tiny $\pm .3$} & \cellcolor{inndomain}56.1{\tiny $\pm .1$} & \cellcolor{indomain}65.2{\tiny $\pm .§$} \\
& \texttt{+ Self-refine} & \cellcolor{outdomain}57.6{\tiny $\pm .3$} & \cellcolor{outdomain}58.7{\tiny $\pm .4$} & \cellcolor{outdomain}59.3{\tiny $\pm .2$} & \cellcolor{outdomain}51.4{\tiny $\pm .2$} & \cellcolor{inndomain}\textbf{63.4}{\tiny $\pm .1$} & \cellcolor{indomain}60.3{\tiny $\pm .1$} \\
& \texttt{Cross Self-refine} & 61.3{\tiny $\pm .5$} & 64.3{\tiny $\pm .2$} & 70.1{\tiny $\pm .4$} & 58.2{\tiny $\pm .1$} & - & 70.5{\tiny $\pm .3$} \\
\hline

\multirow{3}{*}{\textbf{MultiArith}} & \texttt{Instruction-tuning} & \cellcolor{outdomain}57.9{\tiny $\pm .2$} & \cellcolor{outdomain}59.2{\tiny $\pm .3$} & \cellcolor{outdomain}53.8{\tiny $\pm .4$} & \cellcolor{outdomain}51.5{\tiny $\pm .3$} & \cellcolor{indomain}69.3{\tiny $\pm .2$} & \cellcolor{inndomain}89.6{\tiny $\pm .4$} \\
& \texttt{+ Self-refine} & \cellcolor{outdomain}59.1{\tiny $\pm .2$} & \cellcolor{outdomain}63.2{\tiny $\pm .4$} & \cellcolor{outdomain}59.4{\tiny $\pm .5$} & \cellcolor{outdomain}59.9{\tiny $\pm .2$} & \cellcolor{indomain}68.2{\tiny $\pm .1$} & \cellcolor{inndomain}\textbf{91.4}{\tiny $\pm .3$} \\
& \texttt{Cross Self-refine} & 64.7{\tiny $\pm .4$} & 65.8{\tiny $\pm .2$} & 64.1{\tiny $\pm .4$} & 61.5{\tiny $\pm .4$} & 70.1{\tiny $\pm .3$} & - \\
\bottomrule
\end{tabular}
\caption{Evaluation of Mistral-7 Instruction-tuned (\texttt{Instruction-tuned}) and with completely Self-refine Instruction-tuning (\texttt{+ Self-refine Instruction-tuned}) on Demonstrations using different test sets. We evaluate in-domain (QA vs QA) and out-domain (QA vs math-word problem) benchmarks. "Baselines" are referred to the non-instructed model. Results colored in green indicate the in-domain benchmark, blue the out-domain benchmark, and orange the same benchmark on which perform the evaluation phase. Moreover, we propose Self-refine Instruction-tuning in cross-setting scenario where we optimize the model on the training set related to the evaluated task.}
\label{tab:generalization_results_mistral}
\end{table*}

\begin{table*}
\section{Quality of Generations}
\label{app:quality}
To demonstrate the quality of the demonstrations generated by the teachers and students, we propose annotating the responses provided by the teacher and student models automatically. In particular, we sampled 300 questions (50 questions for each task from the testing set split). Hence, we systematically prompt both the teacher LLMs and students. Finally, we estimated the quality of the responses generated by systematically prompting a judge LLM (we chose \texttt{GPT-4} as it is not among the models used in this work). 

\vspace{1cm}

\begin{tabular}{|p{0.9\linewidth}|}
\hline
\texttt{Please act as an impartial judge and evaluate the quality of the response provided by an AI assistant to the user instruction displayed below. Your evaluation should consider factors such as quality, accuracy, depth, and level of detail. Begin your assessment with a short explanation. Be as objective as possible. After providing your explanation, please rate the response on a scale of 1 to 3 strictly following this format:“[[rating]]”, for example: “Rating: [[2]]”.} \\ 
\texttt{[question]} \\ 
\texttt{\$\{question\}} \\ 
\texttt{[AI assistant’s response]} \\ 
\texttt{\$\{response\}} \\ 
\hline
\end{tabular}
\label{tab:prompt_for_quality}
\caption{Using this prompt, we systematically query GPT-4 to note the answers' quality.}

\end{table*}

\begin{table*}

\begin{center}
\begin{tabular}{l|c|c|c}
\hline
\hline
 \textbf{Model} & \textbf{Llama2-70b} & \textbf{Mixtral8x7b} & \textbf{GPT-3.5} \\
\hline
 Baseline & 1.63 & 1.34 & 1.68 \\
 Baseline CoT & 2.72 & 2.56 &\textbf{ 2.89} \\
 Target Answers & 1 & 1 & 1 \\
\hline
\end{tabular}
\end{center}
\label{tab:quality_LLMs}
\caption{Averages quality scores obtained by LLMs' answers by using GPT-4 as judge (see Table \ref{tab:prompt_for_quality}).}

\vspace{1cm}

\begin{center}
\begin{tabular}{l|l|c|c|c}
\hline
\hline
& \textbf{Model} & \textbf{Llama2-7b} & \textbf{Llama2-13b} & \textbf{Mistral-7b} \\
\hline
& Baseline & 1.26 & 1.39 & 1.16 \\
& Baseline CoT & 1.47 & 1.56 & 1.21 \\ 
\hline
\multirow{2}{*}{\textbf{\texttt{in-family}}}
& Instruction-tuning & 2.43 & 2.66 & 2.36 \\
& Self-refine Instruction-tuning & 2.75 & \textbf{2.83} & 2.54 \\
\hline
\multirow{2}{*}{\textbf{\texttt{out-family (GPT-3.5)}}}
& Instruction-tuning & 1.99 & 2.17 & 1.76 \\
& Self-refine Instruction-tuning & \textbf{2.86} & 2.79 & \textbf{2.82} \\
\hline
\end{tabular}
\end{center}
\label{tab:quality_student}
\caption{Averages quality scores obtained by students' answers by using GPT-4 as judge (see Table \ref{tab:prompt_for_quality}).}

\end{table*}

\begin{table*}
\small
    \begin{center}
\begin{tabular}{l|c|c|c|c}
\hline
\hline
 \textbf{work} & \textbf{approach} & \textbf{teacher/s} & \textbf{students/s} & \textbf{tasks} \\ \hline
 \cite{zelikman2022star} & Self-SFT & - & GPT-J, LaMDA & GSM8k, CSQA \\ \hline
\cite{magister-etal-2023-teaching}  & SFT & PaLM & T5-small, -medium & GSM8k, StrategyQA, \\
  &  & GPT-3.5 & T5-large, -xxl & MArith \\ \hline

\cite{li-etal-2023-symbolic} & SFT & GPT-3 175B & OPT-1.3b & CSQA, OBQA, QARel \\
 &  & & & \\ \hline

\cite{shridhar-etal-2023-distilling} &  SFT & GPT-3 175B & GPT-2 & GSM8k, StrategyQA \\
 &   & & & SVAMP \\
\hline
\cite{ho-etal-2023-large} &  SFT & InstructGPT & GPT-3 & GSM8k, StrategyQA, MArith, \\
 &   & (text-davinci-002) &  (ada,babbage,curie) & SVAMP, AddSub \\ \hline

 \cite{wang-etal-2023-democratizing} & IT+RL & GPT-3 & GPT-J & GSM8K, MultiArith, SVAMP \\ 
   &  &  &  &  CSQA, StrategyQA \\\hline
 \cite{luong2024reft} & SFT+RL & GPT-3.5 & Galactica, CodeLlama & GSM8k SVAMP MathQA \\ \hline
 \cite{ranaldi-freitas-2024-aligning} & IT & GPT-3.5, Llama2-70 & Llama2-7,13, Mistral-7 & GSM8k, PIQA, MathQA \\ 
    &  &  &  &  CSQA, OBQA, SIQA \\\hline
 \cite{wang2023making} & SFT+RL & GPT-3.5 & Llama2-7,13 & GSM8k, EAQA \\ \hline
\cite{paul2024refiner} & SFT & GPT-3.5 & CodeT5 s,m & GSM8k, SVAMP, MArith \\ \hline

 & IT+RL (DPO) & GPT-3.5, Llama2-70 & Llama2-7,Llama2-13, & GSM8k, CSQA, OBQA \\ 
\textbf{Ours} & (in-family vs out-family) & Mixtral8x7 & Mistral-7 & PIQA, SIQA, MArith \\ 
  &  & & & MATH, MMLU \\ \hline

\hline
\hline
\end{tabular}
\end{center}
\caption{Summary of methods, teacher and student models of previous work, we indicate Supervised Fine-tuning as (SFT), Instruction-tuning as (IT), and Reinforcement Learning (RL). *note that previous works do not use DPO \cite{rafailov2023direct}}
\label{tab:resume_rel_work}
\end{table*}

\begin{figure*}[t]
\section{Additional Evaluations}
\label{app:additionals}
\centering
         \begin{minipage}{0.3\linewidth}
     \centering
     \includegraphics[width=\linewidth]{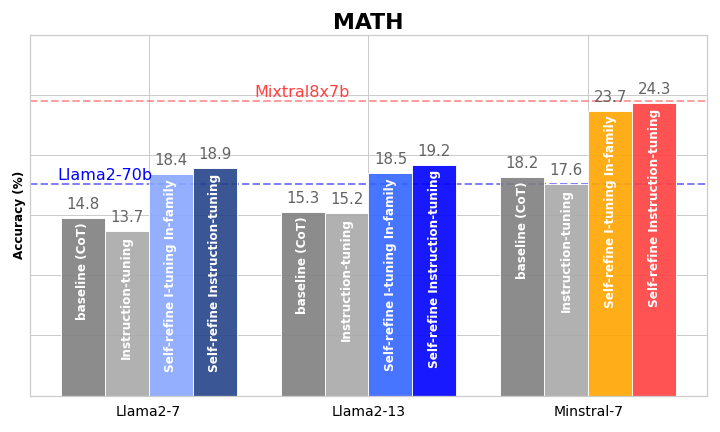}
   \end{minipage}
            \begin{minipage}{0.3\linewidth}
     \centering
     \includegraphics[width=\linewidth]{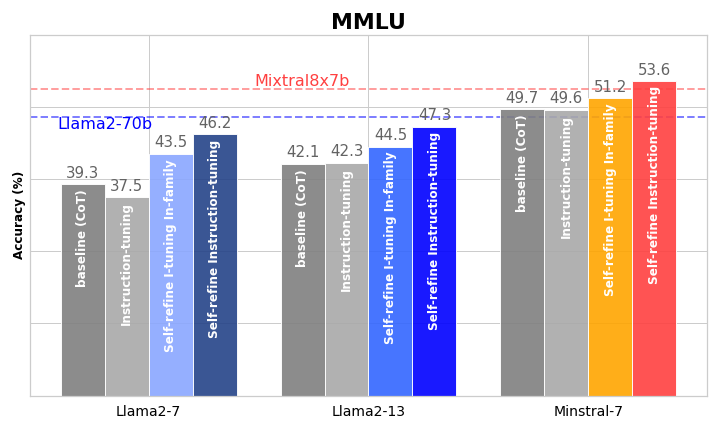}
   \end{minipage}

        \begin{minipage}{0.64\linewidth}
     \centering
     \includegraphics[width=\linewidth]{img/in-family/legend_in_family.png}
   \end{minipage}

            \begin{minipage}{0.3\linewidth}
     \centering
     \includegraphics[width=\linewidth]{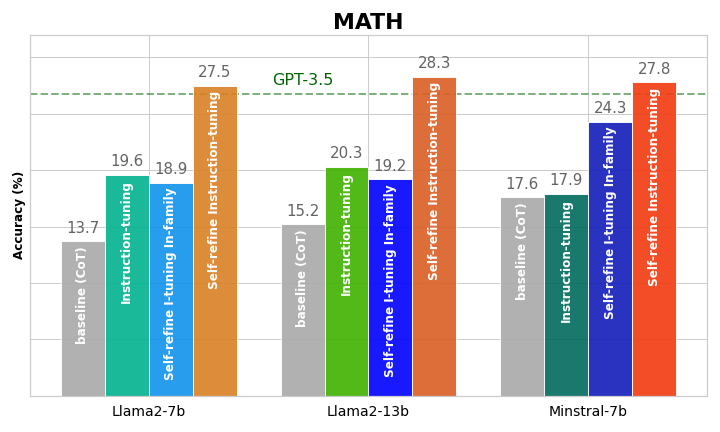}
   \end{minipage}
   %\hfill
            \begin{minipage}{0.3\linewidth}
     \centering
     \includegraphics[width=\linewidth]{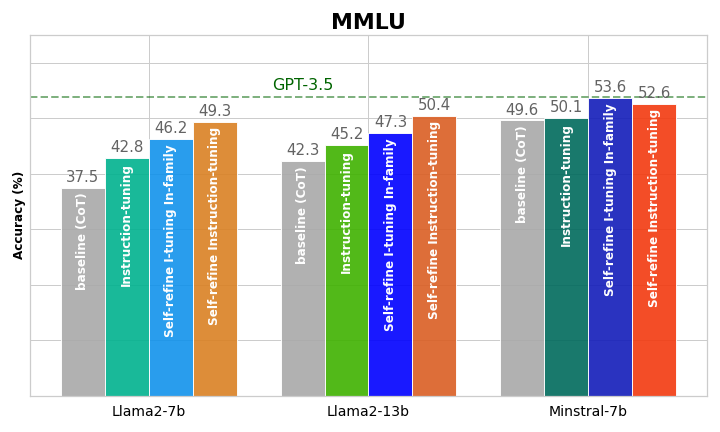}
   \end{minipage}

        \begin{minipage}{0.64\linewidth}
     \centering
     \includegraphics[width=\linewidth]{img/out-family/legend_out_family.png}
   \end{minipage}

   \caption{Accuracies (\%) additional benchmarks as described in Section \ref{sec:data}. Applying the same pipeline proposed in Section \ref{sec:method} and the same experimental set-up (Section \ref{sec:Experimental_Setup}) as the experiments shown in Figure \ref{fig:performances_in_family} and Figure \ref{fig:performances_out_family}. In this experiment, we showed that the approach proposed in Section \ref{sec:method} is also scalable on multi-task benchmarks such as MATH \cite{hendrycks2021measuring} and MMLU \cite{hendrycks2020measuring}. (Self-refine Instruction-tuning phase performed using 25\% as the training set and omitted in the evaluation phase) (as described in the legend, we use the notation \textit{method(Teacher->Student)}). } 
   \label{fig:performances_others_benckmarks}

\end{figure*}

\end{document}